\documentclass{article}

\usepackage[utf8]{inputenc} 
\usepackage[T1]{fontenc}    
\usepackage{hyperref}       
\usepackage{url}            
\usepackage{booktabs}       
\usepackage{amsfonts}       
\usepackage{nicefrac}       
\usepackage{microtype}      
\usepackage{xcolor}         

\usepackage{wrapfig}
\usepackage{graphicx}
\usepackage{amsmath}        
\usepackage{amsthm}         
\usepackage{mathtools}

\usepackage{multirow}       
\usepackage{multicol}       

\usepackage[lofdepth,lotdepth]{subfig}      

\usepackage[acronym, section=section, nonumberlist, nomain, nopostdot]{glossaries}  

\usepackage{natbib}
\setcitestyle{round}

\usepackage[final]{neurips_2022}

\title{Generalised Mutual Information\\for Discriminative Clustering}

\author{%
  Louis Ohl\\
  Universit\'e C\^ote d'Azur\\ Inria, CNRS\\ I3S, Maasai team\\
  CHU de Qu\'ebec Research Center\\Laval University\\\texttt{louis.ohl@inria.fr}
  \And
  Pierre-Alexandre Mattei\\
  Universit\'e C\^ote d'Azur\\ Inria, CNRS \\ LJAD, Maasai team\\\texttt{pierre-alexandre.mattei@inria.fr}
  \AND
  Charles Bouveryon\\
  Universit\'e C\^ote d'Azur\\ Inria, CNRS\\ LJAD, Maasai team\\
  \And
  Warith Harchaoui\\
  Jellysmack\\ AI Labs \\ Research and Development 
  \And
  Mickael Leclerq\\
  CHU de Qu\'ebec Research Center\\Laval University\\
  \AND
  Arnaud Droit\\
  CHU de Qu\'ebec Research Center\\Laval University\\
  \And
  Frederic Precioso\\
  Universit\'e C\^ote d'Azur\\ Inria, CNRS\\ I3S, Maasai team
}

\renewcommand{\vec}{\pmb}

\newcommand{\p}{p_\theta}

\newcommand{\E}{\mathbb{E}}
\newcommand{\I}{\mathcal{I}}

\newcommand{\px}{p_\textup{data}(\vec{x})}
\newcommand{\ya}{y_a}
\newcommand{\yb}{y_b}
\newcommand{\py}{\p(y)}
\newcommand{\pya}{\p(\ya)}
\newcommand{\pyb}{\p(\yb)}
\newcommand{\pyx}{\p(y|\vec{x})}
\newcommand{\pyxa}{\p(\ya|\vec{x})}
\newcommand{\pyxb}{\p(\yb|\vec{x})}
\newcommand{\pxy}{\p(\vec{x}|y)}
\newcommand{\pxya}{\p(\vec{x}|\ya)}
\newcommand{\pxyb}{\p(\vec{x}|\yb)}
\newcommand{\xa}{\vec{x}_a}

\newcommand{\xb}{\vec{x}_b}

\newcommand{\PX}{p(x)}
\newcommand{\PY}{p(y)}

\newcommand{\PXY}{p(x|y)}

\newcommand{\Pjoint}{p(x,y)}

\renewcommand{\cite}{\citep}

\newacronym{gemini}{GEMINI}{generalised mutual information}
\newacronym{kl}{KL}{Kullback-Leibler}
\newacronym{mi}{MI}{mutual information}
\newacronym{ovo}{OvO}{\emph{one-vs-one}}
\newacronym{ova}{OvA}{\emph{one-vs-all}}
\newacronym{mmd}{MMD}{maximum mean discrepancy}
\newacronym{mlp}{MLP}{multi-layered perceptron}
\newacronym{ipm}{IPM}{integral probability metrics}
\newacronym{gmm}{GMM}{gaussian mixture model}
\newacronym{tv}{TV}{total variation}
\newacronym{ari}{ARI}{adjusted rand index}

\begin{document}

\maketitle
\begin{abstract}
  In the last decade, recent successes in deep clustering majorly involved the \acrfull*{mi} as an unsupervised objective for training neural networks with increasing regularisations. While the quality of the regularisations have been largely discussed for improvements, little attention has been dedicated to the relevance of \acrshort*{mi} as a clustering objective. In this paper, we first highlight how the maximisation of \acrshort*{mi} does not lead to satisfying clusters. We identified the Kullback-Leibler divergence as the main reason of this behaviour. Hence, we generalise the mutual information by changing its core distance, introducing the \acrfull*{gemini}: a set of metrics for unsupervised neural network training. Unlike \acrshort*{mi}, some GEMINIs do not require regularisations when training. Some of these metrics are geometry-aware thanks to distances or kernels in the data space. Finally, we highlight that GEMINIs can automatically select a relevant number of clusters, a property that has been little studied in deep clustering context where the number of clusters is a priori unknown.
\end{abstract}

\section{Introduction}
\label{sec:introduction}
Clustering is a fundamental learning task which consists in separating data samples into several categories, each named cluster. This task hinges on two main questions concerning the assessment of correct clustering and the actual number of clusters that may be contained within the data distribution. However, this problem is ill-posed since a cluster lacks formal definitions which makes it a hard problem \cite{kleinberg_impossibility_2003}.

Model-based algorithms make assumptions about the true distribution of the data as a result of some latent distribution of clusters \cite{bouveyron_model_2019}. These techniques are able to find the most likely cluster assignment to data points. These models are usually generative, exhibiting an explicit assumption of the prior knowledge on the data.

Early deep models to perform clustering first relied on autoencoders, based on the belief that an encoding space holds satisfactory properties \cite{xie_unsupervised_2016,ghasedi_dizaji_deep_2017,ji_invariant_2019}. However, the drawback of these architectures is that they do not guarantee that data samples which should meaningfully be far apart remain so in the feature space. Early models that dropped decoders notably used the \gls*{mi} \cite{krause_discriminative_2010,hu_learning_2017} as an objective to maximise. The \gls*{mi} can be written in two ways, either as measure of dependency between two variables $x$ and $y$, e.g. data distribution $\PX$ and cluster assignment $\PY$:
\begin{equation}
\label{eq:mutual_information_1}
\I(x;y) = D_\text{KL} (\Pjoint || \PX\PY),
\end{equation}
or as an expected distance between implied distributions and the overall data:
\begin{equation}
\label{eq:mutual_information_2}
\I(x;y)= \mathbb{E}_{\PY} [ D_\text{KL} (\PXY||\PX)],
\end{equation}
with $D_\text{KL}$ being the \gls*{kl} divergence. Related works often relied on the notion of \gls*{mi} as a measure of coherence between cluster assignments and data distribution \cite{hjelm_learning_2019}. Regularisation techniques were employed to leverage the potential of \gls*{mi}, mostly by specifying model invariances, for example with data augmentation \cite{ji_invariant_2019}.

The maximisation of \gls*{mi} thus gave way to contrastive learning objectives which aim at learning stable representations of data through such invariance specifications \cite{chen_simple_2020,caron_unsupervised_2020}. The contrastive loss maximises the similarity between the features of a sample and its augmentation, while decreasing the similarity with any other sample. Clustering methods also benefited from recent successful deep architectures \cite{li_contrastive_2021,tao_clustering-friendly_2021,huang_deep_2020} by encompassing regularisations in the architecture. These methods correspond to discriminative clustering where we seek to directly infer cluster given the data distribution.
Initial methods also focused on alternate schemes, for example with curriculum learning \cite{chang_deep_2017} to iteratively select relevant data samples for training: for example by alternating K-means cluster assignment with supervised learning using the inferred labels \cite{caron_deep_2018}, or by proceeding to multiple distinct training steps \cite{van_gansbeke_scan_2020,dang_nearest_2021,park_improving_2021}.

However, most of the methods above rarely discuss their robustness when the number of clusters to find is different from the amount of preexisting known classes. While previous work was essentially motivated by considering \gls*{mi} as a dependence measure, we explore in this paper the alternative definition of the \gls*{mi} as the expected distance between data distribution implied by the clusters and the entire data. We extend it to incorporate cluster-wise comparisons of implied distributions, and question the choice of the \gls*{kl} divergence with other possible statistical distances.

Throughout the introduction of the \gls*{gemini}, the contributions of this paper are:

\begin{itemize}
\item A demonstration of how the maxima of \gls*{mi} are not sufficient criteria for clustering. This extends the contribution of \cite{tschannen_mutual_2019} to the discrete case.
\item The introduction of a set of metrics called \glspl*{gemini} involving different distances between distributions which can incorporate prior knowledge on the geometry of the data. Some of these metrics do not require regularisations.
\item A highlight of the implicit selection of clusters from \glspl*{gemini} which allows to select a relevant number of cluster during training.
\end{itemize}

\section{Is mutual information a good clustering objective?}
\label{sec:about_mi}
We consider in this section a dataset consisting in $N$ unlabelled samples $\mathcal{D}=\{\vec{x}_i\}_{i=1}^N$. We distinguish two major use cases of the mutual information: one where we measure the dependence between two continuous variables, as is the case in representation learning, and one where the random variable is discrete. In representation learning, the goal is to construct a continuous representation $\vec{z}$ extracted from the data $\vec{x}$ using a learnable distribution of parameters $\theta$. In clustering, samples $\vec{x}$ are assigned to the discrete variable $y$ through another learnable distribution.


\subsection{Representation learning}
\label{ssec:mi_representation}

Representation learning consists in finding high-level features $\vec{z}_i$ extracted from the data $\vec{x}_i$ in order to perform a \emph{downstream task}, e.g. clustering or classification. \gls*{mi} between $\vec{x}$ and $\vec{z}$ is a common choice for learning features\cite{hjelm_learning_2019}. However, estimating correctly \gls*{mi} between two random variables in continuous domains is often intractable when $p(\vec{x}|\vec{z})$ or $p(\vec{z}|\vec{x})$ is unknown, thus lower bounds are preferred, e.g. variational estimators such as MINE \cite{belghazi_mine_2018}, $\I_\text{NCE}$\cite{van_den_oord_representation_2018}. Another common choice of loss function to train features are contrastive losses such as NT-XENT \cite{chen_simple_2020} where the similarity between the features $\vec{z}_i$ from data $\vec{x}_i$ is maximised with the features $\tilde{\vec{z}}$ from a data-augmented $\tilde{\vec{x}}_i$ against any other features $\vec{z}_j$. Recently, \citet{do_clustering_2021} achieved excellent performances in single-stage methods by highlighting the link between the $\I_\text{NCE}$ estimator \cite{van_den_oord_representation_2018} and contrastive learning losses. Representation learning therefore comes at the cost of a complex lower bound estimator on \gls*{mi}, which often requires data augmentation.
Moreover, it was noticed that the \gls*{mi} is hardly predictive of downstream tasks \cite{tschannen_mutual_2019} when the variable $y$ is continuous, i.e. a high value of \gls*{mi} does not clarify whether the discovered representations are insightful with regards to the target of the downstream task.

\subsection{Discriminative clustering}
\label{sssec:mi_clustering}


The \gls*{mi} has been first used as an objective for learning discriminative clustering models \cite{bridle_unsupervised_1992}. Associated architectures went from simple logistic regression \cite{krause_discriminative_2010} to deeper architectures \cite{hu_learning_2017,ji_invariant_2019}. Beyond architecture improvement, the \gls*{mi} maximisation was also carried with several regularisations. These regularisations include penalty terms such as weight decay \cite{krause_discriminative_2010} or Virtual Adversarial Training (VAT, \citealp{hu_learning_2017,miyato_virtual_2018}). Data augmentation was further used to provide invariances in clustering, as well as specific architecture designs like auxiliary clustering heads \cite{ji_invariant_2019}. Rewriting the \gls*{mi} in terms of entropies:
\begin{equation}
\label{eq:mi_kl_entropies}
\I (\vec{x};y) = \mathcal{H}(y) - \mathcal{H}(y|\vec{x})
\end{equation}
highlights a requirement for balanced clusters, through the cluster entropy term $\mathcal{H}(y)$. Indeed, a uniform distribution maximises the entropy. This hints that an unregularised discrete mutual information for clustering can possibly produce uniformly distributed clusters among samples, regardless of how close they could be. We highlight this claim in section~\ref{ssec:local_maxima}. As an example of regularisation impact: maximising the \gls*{mi} with $\ell_2$ constraint can be equivalent to a soft and regularised K-Means in a feature space \cite{jabi_deep_2019}. In clustering, the number of clusters to find is usually not known in advance. Therefore, an interesting clustering algorithm should be able to find a relevant number of clusters, i.e. perform model selection. However, model selection for parametric deep clustering models is expensive \cite{ronen_deepdpm_2022}. Cluster selection through \gls*{mi} maximisation has been little studied in related works, since experiments usually tasked models to find the (supervised) classes of datasets. Furthermore, the literature diverged towards deep learning methods focusing mainly on images, yet rarely on other type of data such as tabular data \cite{min_survey_2018}.

\subsection{Maximising the MI can lead to bad decision boundaries}
\label{ssec:local_maxima}

\begin{figure}[t]
    \centering
    \subfloat[Good decision boundary]{
        \includegraphics[height=0.1\paperheight]{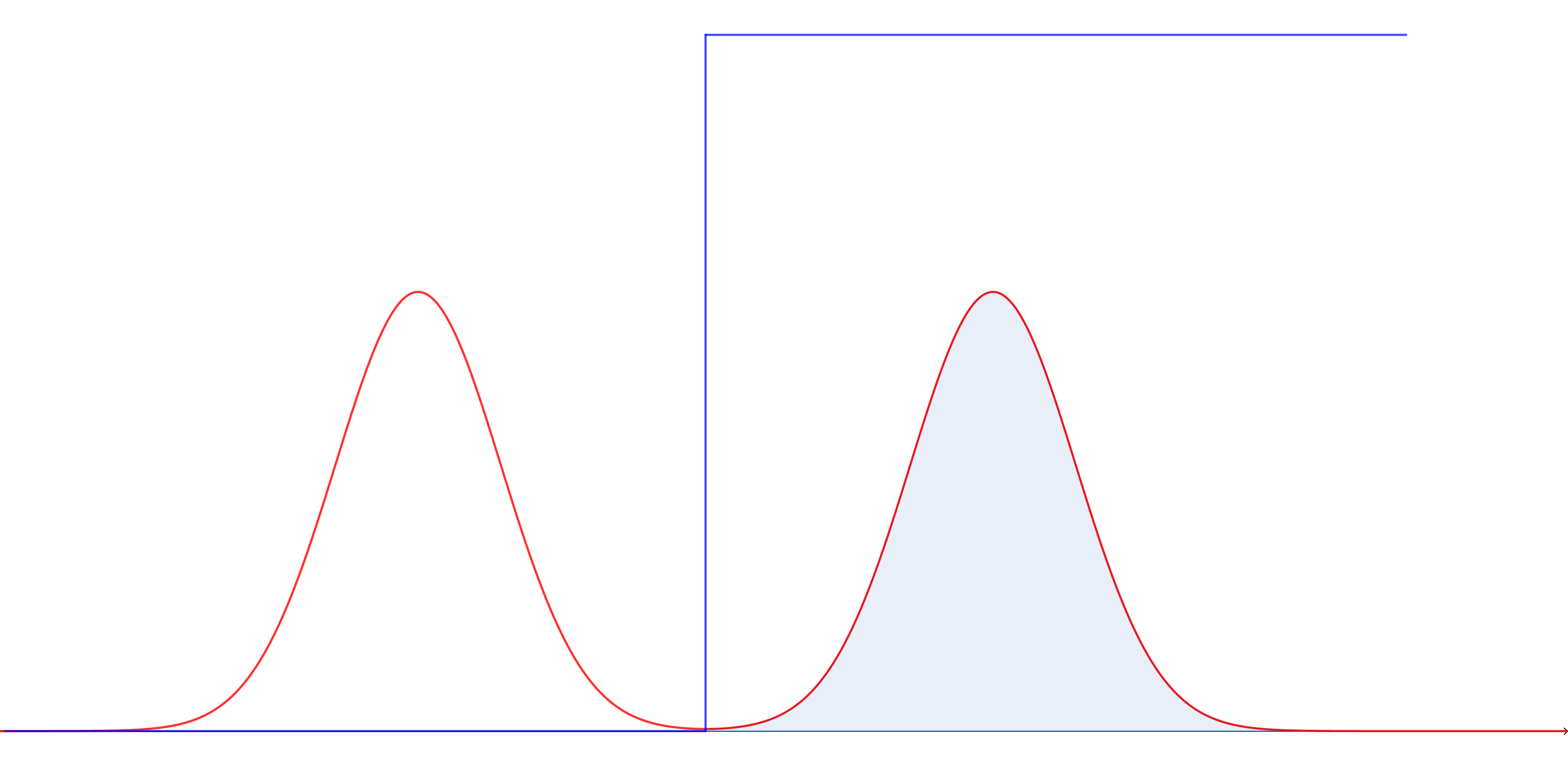}
        \label{sfig:good_decision_boundary}
    }\hfill
    \subfloat[Misplaced decision boundary]{
        \includegraphics[height=0.1\paperheight]{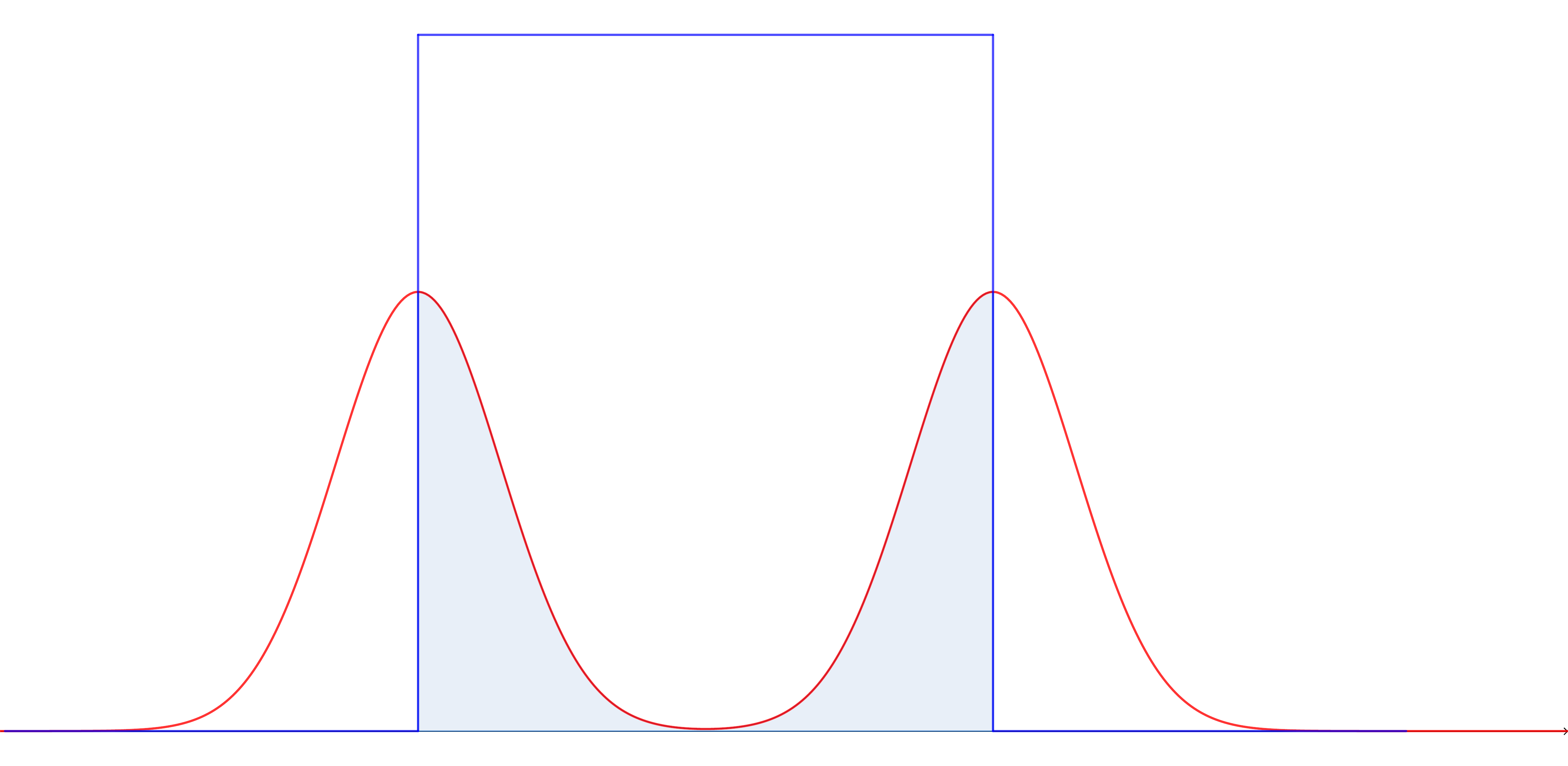}
        \label{sfig:odd_decision_boundary}
    }
    \caption{Example of maximised MI for a Gaussian mixture $p(x) \sim \frac{1}{2} \mathcal{N}(\mu_0, \sigma^2)+\frac{1}{2}\mathcal{N}(\mu_1,\sigma^2)$. It is clear that Figure~\ref{sfig:good_decision_boundary} presents the best decision boundary and posterior between the two Gaussian distributions. Yet, as $p(x|y)$ converges to a Dirac distribution, the MI difference converges to 0.}
    \label{fig:example_good_odd_mi}
\end{figure}

Maximising the \gls*{mi} directly can be a poor objective: a high \gls*{mi} value is not necessarily predictive of the quality of the features regarding downstream tasks \cite{tschannen_mutual_2019} when $y$ is continuous. We support a similar argument for the case where the data $x$ is a continuous random variable and the cluster assignment $y$ a categorical variable. Indeed, the \gls*{mi} can be maximised by setting appropriately a sharp decision boundary which partitions evenly the data. This reasoning can be seen in the entropy-based formulation of the \gls*{mi} (Eq. \ref{eq:mi_kl_entropies}): any sharp decision boundary minimises the negative conditional entropy, while ensuring balanced clusters maximises the entropy of cluster proportions. Consider for example Figure ~\ref{fig:example_good_odd_mi}, where a mixture of Gaussian distributions with equal variances is separated by a sharp decision boundary. We highlight that both models will have the same mutual information on condition that the misplaced decision boundary of Figure~\ref{sfig:odd_decision_boundary} splits evenly the dataset (see Appendix~\ref{app:mi_convergence}).

Globally, \gls*{mi} misses the idea in clustering that any two points close to one another may be in the same cluster according to some chosen metric. Hence regularisations are required to ensure this constraint. A sketch of these insights was mentionned by \citet{bridle_unsupervised_1992,corduneanu_information_2012}.

\section{Extending the mutual information to the generalised mutual information}
\label{sec:gemini}
Given the identified limitations of MI, we now describe the discriminative clustering framework based on our perception of the mutual information. We then detail the different statistical distances we can use to extend \gls*{mi} to the \acrfull*{gemini}.

\subsection{The discriminative clustering framework for GEMINIs}
\label{ssec:discriminative_clustering}

We change our view on the mutual information by seeing it as a discriminative clustering objective that aims at separating the data distribution given cluster assignments $p(\vec{x}|y)$ from the data distribution $p(\vec{x})$ according to the \gls*{kl} divergence:
\begin{equation}\label{eq:base_mi}
    \I(\vec{x};y) = \E_{y\sim p(y)} \left[ D_\text{KL}(p(\vec{x}|y)\|p(\vec{x}))\right].
\end{equation}
To highlight the discriminative clustering design, we explicitly remove our hypotheses on the data distribution by writing $\px$.  The only part of the model that we design is a conditional distribution $\pyx$ that assigns a cluster $y$ to a sample $\vec{x}$ using the parameters $\theta$ \cite{minka_discriminative_2005}. This conditional distribution can typically be a neural network of adequate design regarding the data, e.g. a CNN, or a simple categorical distribution. Consequently, the cluster proportions are controlled by $\theta$ because $\py=\mathbb{E}[\pyx]$ and so is the conditional distribution $\pxy$ even though intractable. This questions how Eq. (\ref{eq:base_mi}) can be computed. Fortunately, well-known properties of \gls*{mi} can invert the distributions on which the KL divergence is computed \cite{bridle_unsupervised_1992,krause_discriminative_2010} via Bayes' theorem:
\begin{equation} \label{eq:tractable_discriminative_mi}
    \I(\vec{x};y) = \E_{\vec{x} \sim \px} \left[ D_\text{KL} (\pyx \| \py)\right],
\end{equation}
which is possible to estimate. 
Since we highlighted earlier that the KL divergence in the \gls*{mi} can lead to inappropriate decision boundaries, we are interested in replacing it by other distances or divergences. However, changing it in Eq. (\ref{eq:tractable_discriminative_mi}) would focus on the separation of cluster assignments from the cluster proportions which may be irrelevant to the data distribution. We rather alter Eq. (\ref{eq:base_mi}) to clearly show that we separate data distributions given clusters from the entire data distribution because it allows us to take into account the data space geometry.

\subsection{The GEMINI}
\label{ssec:gemini}

\begin{table}
\centering
\caption{Definition of the GEMINI for $f$-divergences, MMD and the Wasserstein distance. We directly write here the equation that can be optimised to train a discriminative model $\pyx$ via stochastic gradient descent since they are expectations over the data.}
\label{tab:all_geminis}
\begin{tabular}{c c}
\toprule
Name&Equation\\\hline\\
KL OvA/MI& $\E_{\px}\left[D_\text{KL}(\pyx \|\py)\right]$\\
KL OvO& $\E_{\px}[D_\text{KL}(\pyx \| \py))+D_\text{KL}(\py \| \pyx))]$\\
\begin{minipage}{0.2\linewidth}\centering Squared Hellinger\\OvA\end{minipage} & $1-\E_{\px}[\E_{\py}[\sqrt{\frac{\pyx}{\py}}]]$\\
\begin{minipage}{0.2\linewidth}\centering Squared Hellinger\\OvO \end{minipage} & $\E_{\px}[\mathbb{V}_{\py}[\sqrt{\frac{\pyx}{\py}}]]$\\
TV OvA& $\E_{\px} [D_\text{TV} (\pyx \| \py) ]$\\
TV OvO& $\frac{1}{2}\E_{\px}[\E_{\ya,\yb\sim\py}[|\frac{ \pyxa }{ \pya } - \frac{\pyxb}{\pyb}|]]$\\
\midrule\\
MMD OvA& $\E_{\py} \left[ \E_{\xa,\xb \sim \px} \left[ k(\xa, \xb) \left( \frac{\p(y|\xa)\p(y|\xb)}{\py^2} + 1 - 2\frac{\p(y|\xa)}{\py}\right) \right]^{\frac{1}{2}}\right]$ \\
MMD OvO& \begin{minipage}{0.7\linewidth}\centering\begin{multline*}\E_{\ya,\yb \sim \py} \left[ \E_{\xa,\xb \sim \px} \left[ k(\xa, \xb) \left( \frac{\p(\ya|\xa) \p(\ya|\xb)}{\pya^2} \right.\right.\right.\\\left.\left.\left. + \frac{\p(\yb |\xa)\p(\yb|\xb)}{\pyb^2} - 2\frac{\p(\ya |\xa)\p(\yb|\xb)}{\pya\pyb}\right) \right]^{\frac{1}{2}}\right] \end{multline*}\end{minipage}\\
Wasserstein OvA&$\mathbb{E}_{\py}\left[\mathcal{W}_c\left(\sum_{i=1}^N m_i^y\delta_{\vec{x}_i},\sum_{i=1}^N \frac{1}{N}\delta_{\vec{x}_i}\right)\right]$\\
Wasserstein OvO&$\mathbb{E}_{\ya,\yb\sim \py}\left[\mathcal{W}_c\left(\sum_{i=1}^N m_i^{\ya}\delta_{\vec{x}_i},\sum_{i=1}^N m_i^{\yb}\delta_{\vec{x}_i}\right)\right]$\\
\bottomrule
\end{tabular}
\end{table}

The goal of the \gls*{gemini} is to separate data distributions according to an arbitrary distance $D$, i.e. changing the \gls*{kl} divergence for another divergence or distance in the \gls*{mi}. Moreover, we question the evaluation of the distance between the distribution of the data given a cluster assumption $\pxy$ and the entire data distribution $\px$. We argue that it is intuitive in clustering to compare the distribution of one cluster against the distribution of \emph{another cluster} rather than the data distribution. This raises the definition of two \glspl*{gemini}, one named \gls*{ova}:
\begin{equation}\label{eq:gemini_ova}
    \I^\text{OvA}_D(\vec{x};y) = \E_{y \sim \py} \left[ D(\pxy\|\px)\right],
\end{equation}
which compares the cluster distributions to the data distribution, and the \gls*{ovo} in which we independently draw cluster assignments $\ya$ and $\yb$ (see App.~\ref{app:ovo_and_ova} for an \gls*{ovo} justification):
\begin{equation}\label{eq:gemini_ovo}
    \I^\text{OvO}_D(\vec{x};y) = \E_{\ya,\yb \sim \py} \left[ D(\pxya \| \pxyb)\right],
\end{equation}
There exists other distances than the KL to measure how far two distributions $p$ and $q$ are one from the other. We can make a clear distinction between two types of distances, Csiszar's $f$-divergences \cite{csiszar_information-type_1967} and \gls*{ipm} \cite{sriperumbudur_integral_2009}. However, unlike $f$-divergences, \gls*{ipm}-derived distances like the Wasserstein distance or the \gls*{mmd}\cite{gneiting_strictly_2007,gretton_kernel_2012} bring knowledge about the data throughout either a distance $c$ or a kernel $\kappa$: these distances are geometry-aware.

\paragraph{$f$-divergence GEMINIs:} These divergences involve a convex function $f:\mathbb{R}^+\rightarrow\mathbb{R}$ such that $f(1)=0$. This function is applied to evaluate the ratio between two distributions $p$ and $q$, as in Eq.~(\ref{eq:f_divergences_definition}):
\begin{equation}
\label{eq:f_divergences_definition}
D_\text{f-div}(p,q) = \E_{\vec{z} \sim q(\vec{z})} \left[ f\left(\frac{p(\vec{z})}{q(\vec{z})}\right)\right].
\end{equation}
We will focus on three $f$-divergences: the \gls*{kl} divergence, the \gls*{tv} distance and the squared Hellinger distance. While the \gls*{kl} divergence is the usual divergence for the \gls*{mi}, the \gls*{tv} and the squared Hellinger distance present different advantages among $f$-divergences. First of all, both of them are bounded between 0 and 1. It is consequently easy to check when any \gls*{gemini} using those is maximised contrarily to the \gls*{mi} that is bounded by the minimum of the entropies of $\vec{x}$ and $y$ \cite{gray_maximum_1977}. When used as distance between data conditional distribution $\pxy$ and data distribution $\px$, we can apply Bayes' theorem in order to get an estimable equation to maximise, which only involves cluster assignment $\pyx$ and marginals $\py$ (see Table~\ref{tab:all_geminis}). 

\paragraph{MMD GEMINIs:} The \gls*{mmd} corresponds to the distance between the respective expected embedding of samples from distribution $p$ and distribution $q$ in a reproducing kernel hilbert space (RKHS) $\mathcal{H}$:
\begin{equation}\label{eq:mmd_definition}
    \text{MMD}(p\|q) = \| \E_{\vec{z} \sim p(\vec{z})} [\varphi(\vec{z})] - \E_{\vec{z}\sim q(\vec{z})} [\varphi(\vec{z})]\|_\mathcal{H},
\end{equation}
where $\varphi$ is the RKHS embedding. To compute this distance we can use the kernel trick \cite{gretton_kernel_2012} by involving the kernel function $\kappa(\vec{a},\vec{b})=\langle\varphi(\vec{a}),\varphi(\vec{b})\rangle$. We use Bayes' theorem to uncover a version of the \gls*{mmd} that can be estimated through Monte Carlo using only the predictions $\pyx$ (see Table~\ref{tab:all_geminis}).
\looseness=-1

\paragraph{Wasserstein GEMINIs:} This distance is an optimal transport distance, defined as: 
\begin{equation}\label{eq:wasserstein_definition}
    \mathcal{W}_c(p,q) = \left(\inf_{\gamma \in \Gamma(p,q)} \E_{\vec{x}, \vec{z} \sim \gamma(\vec{x},\vec{z})}\left[c(\vec{x},\vec{z}) \right]\right),
\end{equation}
where $\Gamma(p,q)$ is the set of all couplings between $p$ and $q$ and $c$ a distance function in $\mathcal{X}$. Computing the Wasserstein distance between two distributions $\p(\vec{x}|y=k_1)$ and $\p(\vec{x}|y=k_2)$ is difficult in our discriminative context because we only have access to a finite set of samples $N$. To achieve the Wasserstein-GEMINI, we instead use approximations of the distributions with weighted sums of Diracs:
\looseness=-1
\begin{equation}\label{eq:dirac_approximation}
\p(\vec{x}|y=k) \approx \sum_{i=1}^N m_i^k \delta_{\vec{x}_i} = p_N^k,\quad\text{with}\quad m_i^k = \frac{\p(y=k|\vec{x}_i)}{\sum_{j=1}^N\p(y=k|\vec{x}_j)},
\end{equation}
where $\delta_{\vec{x}_i}$ is a Dirac located on sample location $\vec{x}_i\in\mathcal{X}$. The Wasserstein-\gls{ova} and -\gls{ovo} applied to Dirac sums are compatible with the \verb+emd2+ function of the Python optimal transport package~\cite{flamary_pot_2021} which gracefully supports automatic differentation (see Appendix~\ref{app:wasserstein_convergence} for convergence to the expectation). All \glspl*{gemini} are summarised in Table~\ref{tab:all_geminis}, (see Appendix~\ref{app:deriving_geminis} for derivations). 
\looseness=-1

\section{Experiments}
\label{sec:experiments}
For all experiments below, we report the \gls*{ari} \cite{hubert_comparing_1985}, a common metric in clustering. This metric is external as it requires labels for evaluation. It ranges from 0, when labels are independent from cluster assignments, to 1, when labels are equivalent to cluster assignments up to permutations. An \gls*{ari} close to 0 is equivalent to the best accuracy when voting constantly for the majority class, e.g. 10\% on a balanced 10-class dataset. Regarding the \gls*{mmd}- and Wasserstein-\glspl*{gemini}, we used by default a linear kernel and the Euclidean distance unless specified otherwise. All discriminative models are trained using the Adam optimiser \cite{kingma_adam_2014}. We estimate a total of 450 hours of GPU consumption. (See Appendix~\ref{app:requirements} for the details of Python packages for experiments and Appendix~\ref{app:cluster_selection} for further experiments regarding model selection). The code is available at \url{https://github.com/oshillou/GEMINI}
\looseness=-1

\subsection{When the MI fails because of the modelling}
\label{ssec:exp_categorical}
\begin{figure}[b]
    \centering
    \subfloat[KL-OvA (MI)]{
        \fbox{ \includegraphics[height=0.1\paperheight,width=0.4\linewidth]{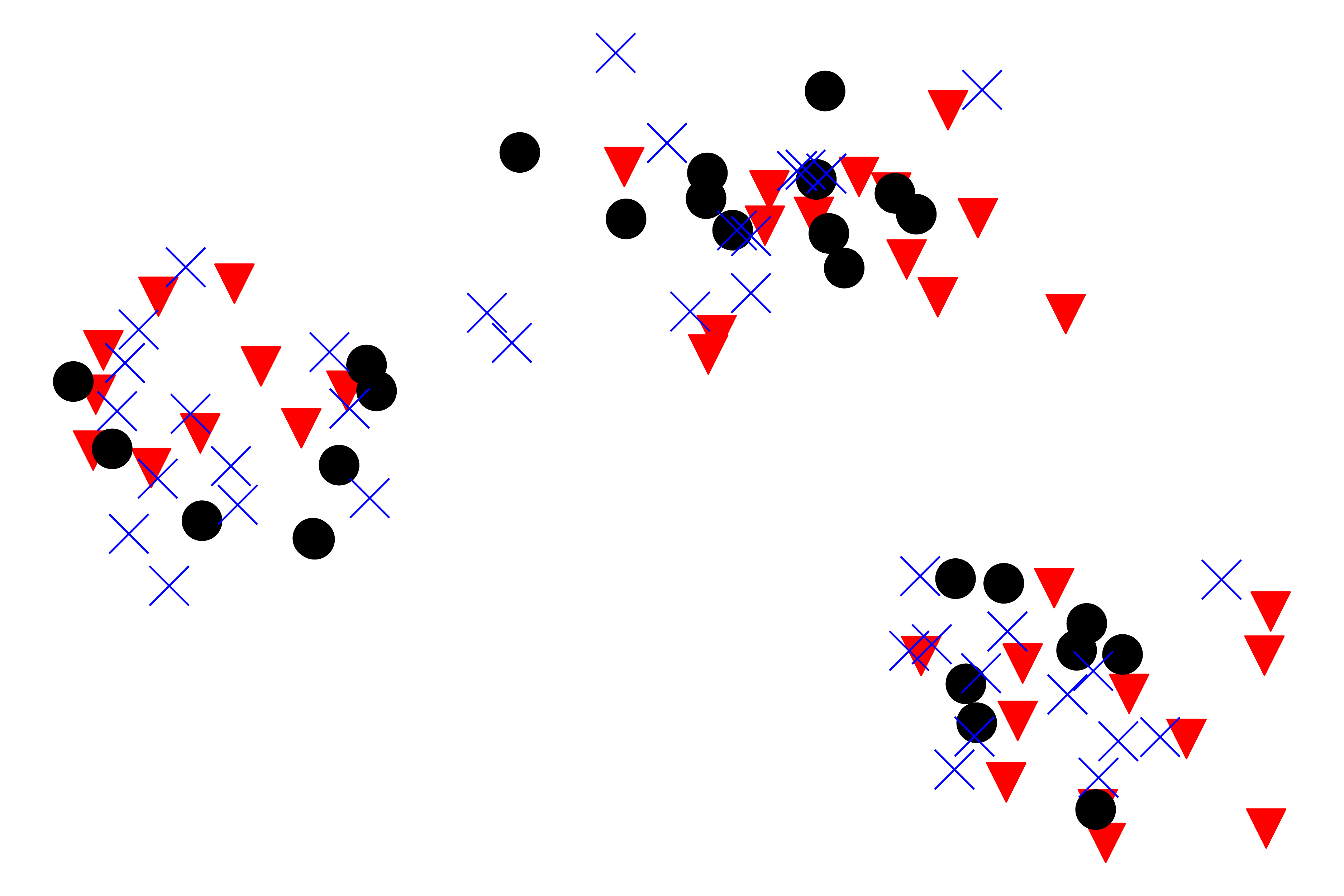}}
        \label{sfig:mi_categorical}
    }
    \subfloat[MMD OvA with linear kernel]{
        \fbox{ \includegraphics[height=0.1\paperheight,width=0.4\linewidth]{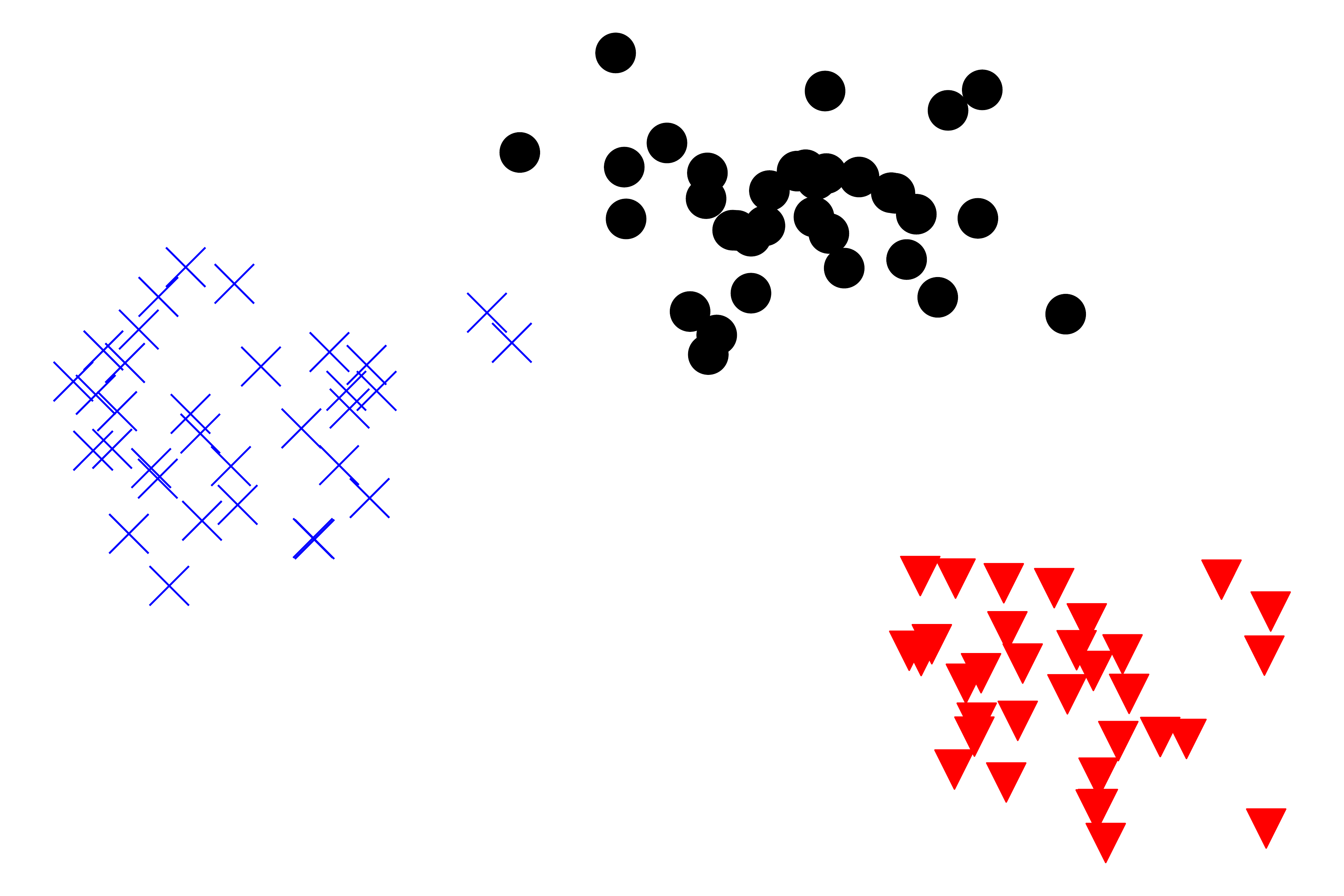}}
        \label{sfig:mmd_categorical}
    }\hfill
    \caption{Clustering of a mixture of 3 Gaussian distributions with MI (left) and a GEMINI (right) using categorical distributions. The MI does not have insights on the data shape because of the model, and clusters points uniformly between the 3 clusters (black dots, red triangles and blue crosses) whereas the MMD is aware of the data shape through its kernel.}
    \label{fig:categorical_boundaries}
\end{figure}

We first took the most simple discriminative clustering model, where each cluster assignment according to the input datum follows a categorical distribution:
\begin{equation*}
    y|\vec{x}=\vec{x}_i \sim \text{Cat}(\theta_i^1,\theta_i^2,\cdots,\theta_i^K).
\end{equation*}
We generated $N=100$ samples from a simple mixture of $K=3$ Gaussian distributions. Each model thus only consists in $NK$ parameters to optimise. This is a simplistic way of describing the most flexible deep neural network. We then maximised on the one hand the \gls*{kl}-\gls*{ova} (\gls*{mi}) and on the other hand the \gls*{mmd}-\gls*{ova}. Both clustering results can be seen in Figure~\ref{fig:categorical_boundaries}. We concluded that without any function, e.g. a neural network, to link the parameters of the conditional distribution with $\vec{x}$, the \gls*{mi} struggles to find the correct decision boundaries. Indeed, the position of $\vec{x}$ in the 2D space plays no role and the decision boundary becomes only relevant with regards to cluster entropy maximisation: a uniform distribution between 3 clusters. However, it plays a major role in the kernel of the \gls*{mmd}-\gls*{gemini} thus solving correctly the problem.

\subsection{Resistance to outliers}
\label{ssec:exp_gstm}

\begin{table}[b]
    \caption{Mean ARI (std) of a MLP fitting a mixture of 3 Gaussian and 1 Student-t multivariate distributions compared with Gaussian Mixture Models and K-Means. The models try to find 4 at best and the Student-t distribution has $\rho$=1 degree of freedom. We write the ARI for the maximum a posteriori of an oracle aware of all parameters of the data.}
    \label{tab:gstm_experiment_ari}
    \centering
    \begin{tabular}{c c c c c c c}
        \toprule
        \multirow{2}{*}{K-Means}&\multicolumn{2}{c}{GMM}&\multicolumn{2}{c}{MMD}&\multicolumn{2}{c}{Wasserstein}\\
        \cmidrule(lr){2-3}\cmidrule(lr){4-5}\cmidrule(lr){6-7}
        &full cov&diagonal cov&$\I_\text{MMD}^\text{ova}$&$\I_\text{MMD}^\text{ovo}$&$\I_\mathcal{W}^\text{ova}$&$\I_\mathcal{W}^\text{ovo}$\\
        \midrule
        0& 0& 0.024& 0.922& 0.921& 0.915&0.922\\
       (0)&(0)&(0.107)&(0.004)&(0.007)&(0.131)&(0.006)\\
        \midrule
        \multirow{2}{*}{Oracle}&\multicolumn{6}{c}{$f$-divergences}\\
        \cmidrule(lr){2-7}
        &$\I_\text{KL}^\text{ova}$&$\I_\text{KL}^\text{ovo}$&$\I_{\text{H}^2}^\text{ova}$&$\I_{\text{H}^2}^\text{ovo}$&$\I_\text{TV}^\text{ova}$&$\I_\text{TV}^\text{ovo}$\\
        \midrule
        \multirow{2}{*}{0.989}&{\bf 0.939}&0.723 &0.906 &0.858 &0.904&{\bf 0.938}\\
        &{\bf (0.006)}&(0.114)& (0.103)& (0.143)& (0.104)&{\bf (0.005)}\\
        \bottomrule
    \end{tabular}
\end{table}
\begin{figure}[hbt]
    \centering
    \subfloat[KL-OvA (MI) Entropy Map]{
        \includegraphics[width=0.3\linewidth]{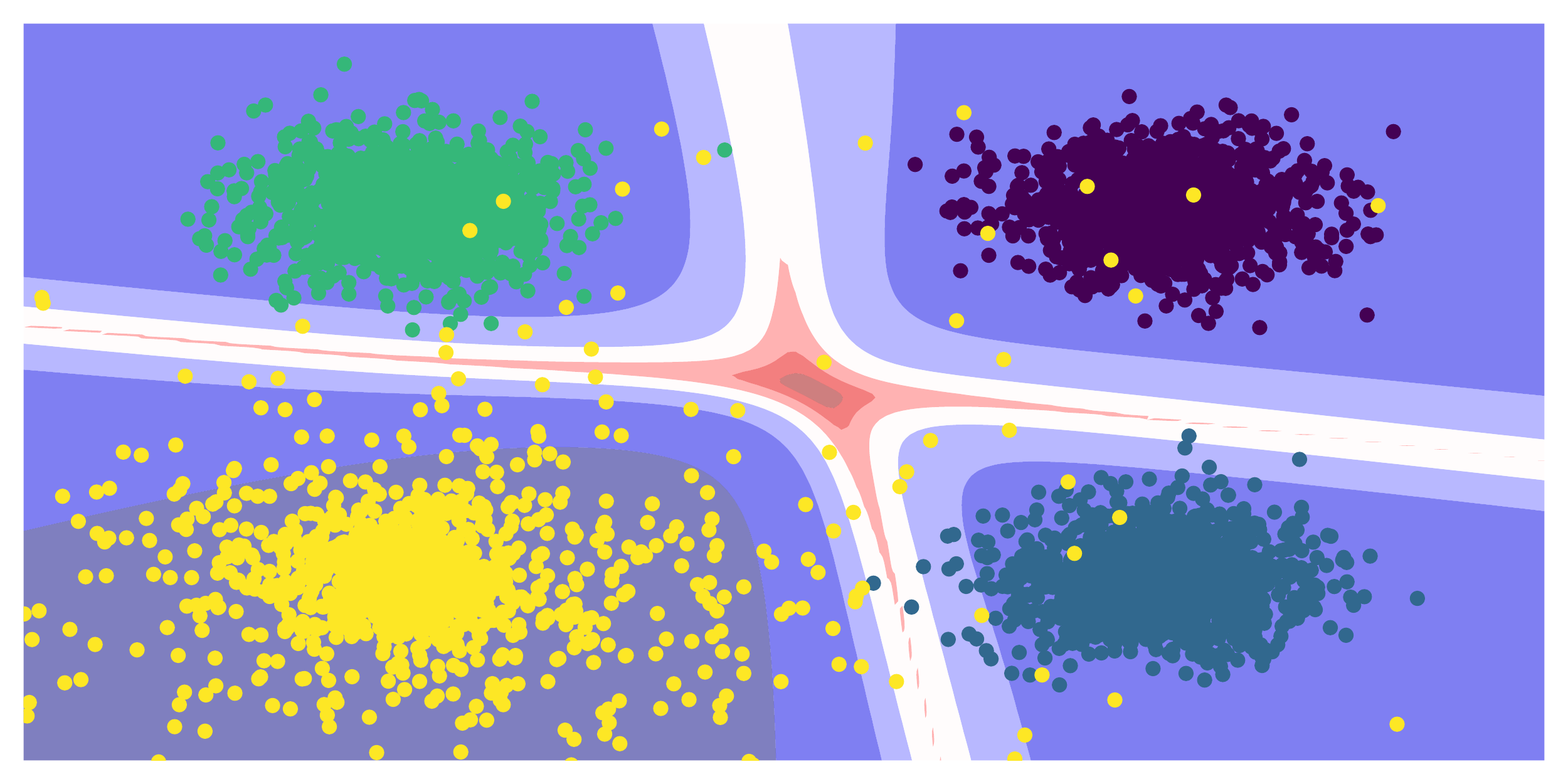}
        \label{sfig:mi_entropy_map}
    }
    \subfloat[MMD-OvO Entropy Map]{
        \includegraphics[width=0.3\linewidth]{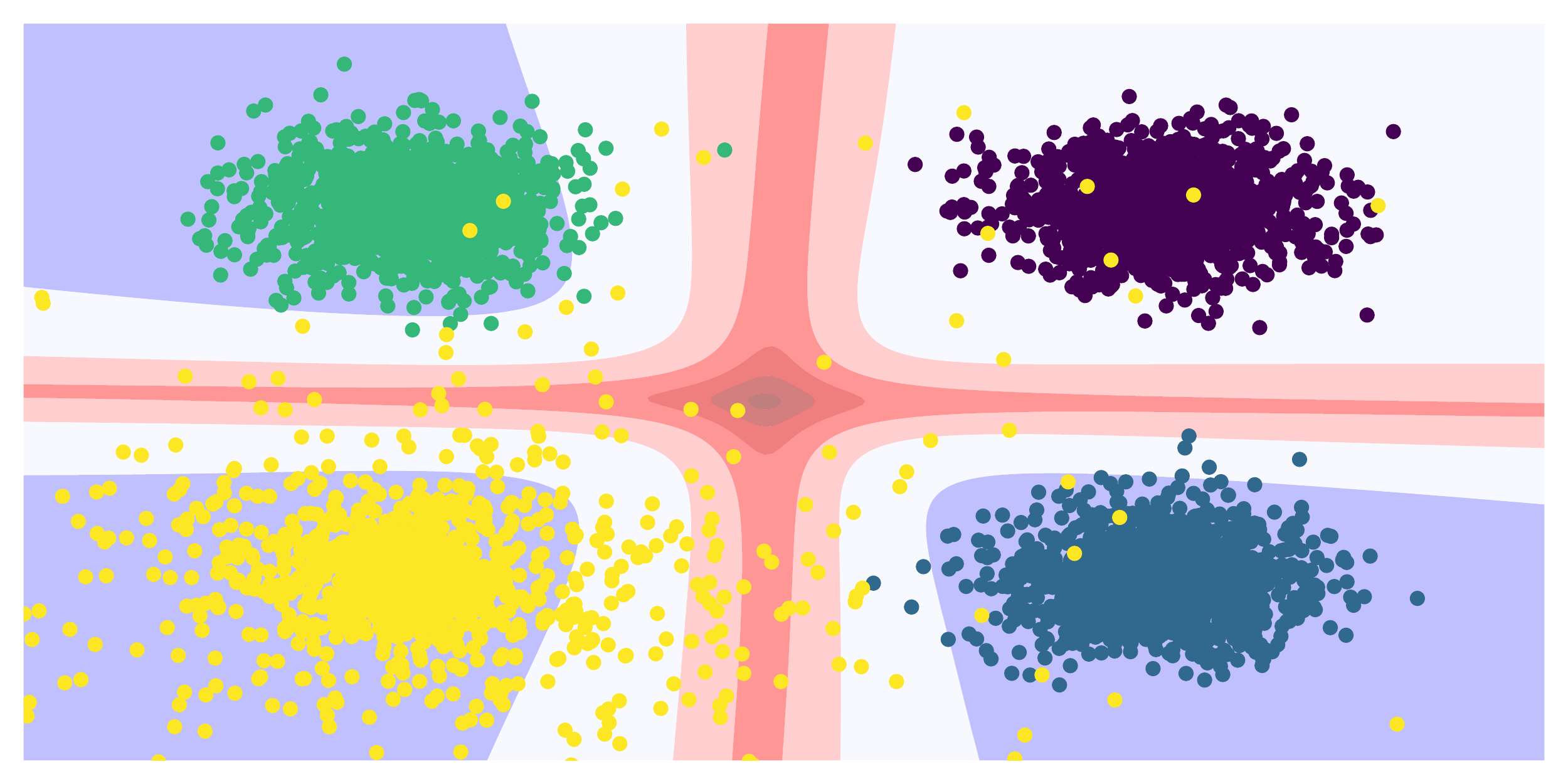}
        \label{sfig:mmd_entropy_map}
    }
    \subfloat[Wasserstein-OvO Entropy Map]{
       \includegraphics[width=0.3\linewidth]{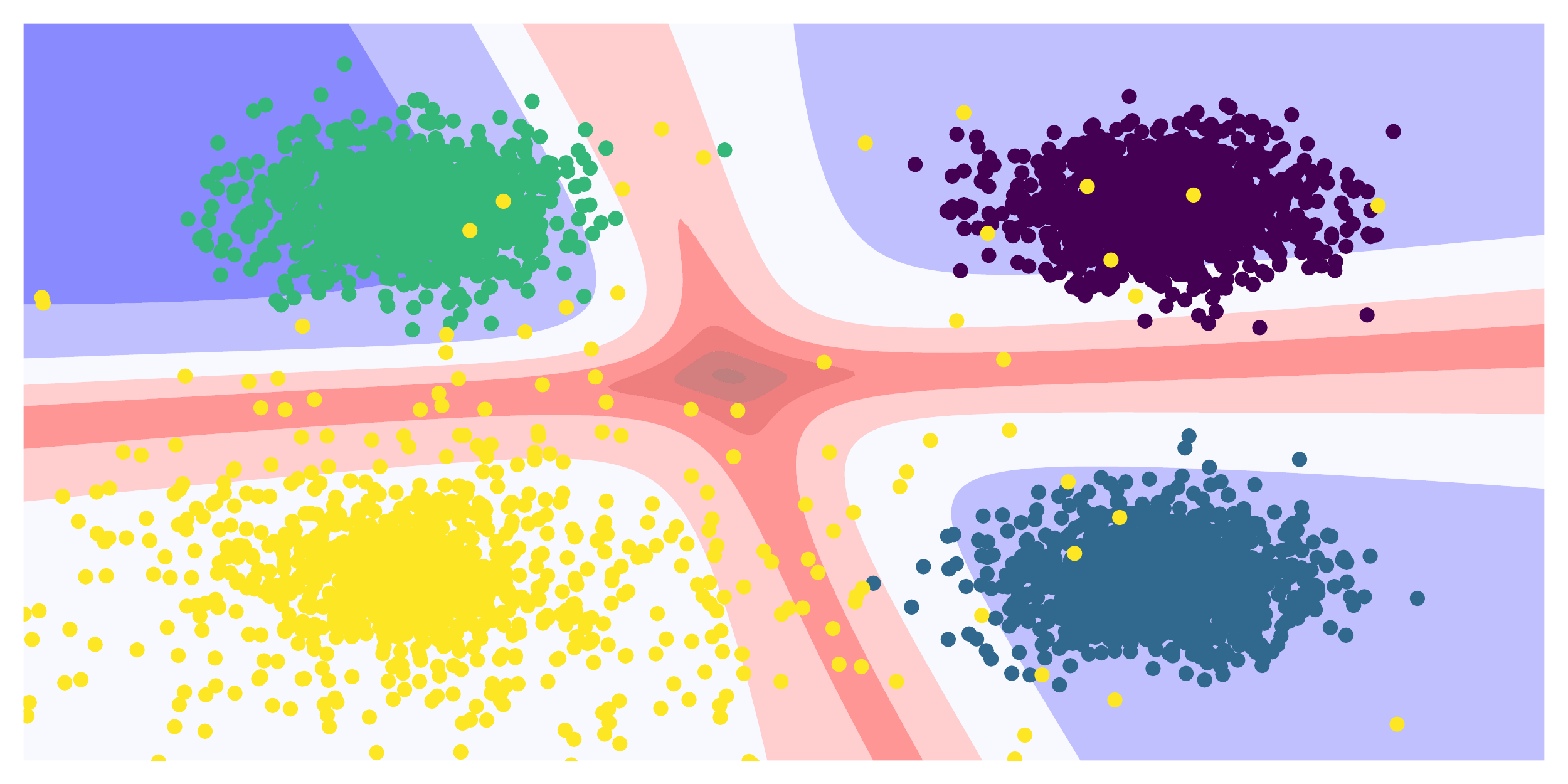}
       \label{sfig:wasserstein_entropy_map}
    }
    \caption{Entropy maps of the predictions of each MLP trained using a GEMINI or the MI. The bottom-left distribution (yellow) is a Student-t distribution with 1 degree of freedom that produces samples far from the origin. The Rényi entropy of prediction is highlighted from lowest (red background) to highest (blue background). MI on the left has the most confident predictions overall and the smallest uncertainty around the decision boundary, i.e. high entropy variations.}
    \label{fig:gstm_entropy_maps}
\end{figure}

To prove the strength of using neural networks for clustering trained with GEMINI, we introduced extreme samples in Gaussian mixtures by replacing a Gaussian distribution with a Student-t distribution for which the degree of freedom $\rho$ is small. We fixed $K=4$ clusters, 3 being drawn from multivariate Gaussian distributions and the last one from a multivariate Student-t distribution in 2 dimensions for visualisation purposes with 1 degree of freedom (see Appendix\ref{app:gstm} for other parameters and results). Thus, the Student-t distribution has an undefined expectation and produces samples that can be perceived as outliers regarding a Gaussian mixture owing to its heavy tail. We report the ARIs of \gls*{mlp} trained 20 times with GEMINIs in Table~\ref{tab:gstm_experiment_ari}. The presence of "outliers" leads K-Means and Gaussian Mixture models to fail at grasping the 4 distributions when tring to find 4 clusters. Meanwhile, \glspl*{gemini} perform better. Note that all \gls*{mmd} and Wasserstein-\gls*{ovo}-\gls*{gemini} present lower standard deviation for high scores compared to $f$-divergence \glspl*{gemini}. We attribute these performances to both the \gls*{mlp} that tries to find separating hyperplanes in the data space and the absence of hypotheses regarding the data. Moreover, as mentioned in section~\ref{ssec:local_maxima}, the usual \gls*{mi} is best maximised when its decision boundary presents little entropy $\mathcal{H}(\pyx)$. As neural networks can be overconfident \cite{guo_calibration_2017}, \gls*{mi} is likely to yield overconfident clustering by minimizing the conditional entropy. We highlight such behaviour in Figure~\ref{fig:gstm_entropy_maps} where the Rényi entropy \cite{renyi_measures_1961} associated to each sample in the \gls*{mi} (Figure~\ref{sfig:mi_entropy_map}) is much lower, if not 0, compared to \gls*{mmd}-\gls*{ovo} and Wasserstein-\gls*{ovo} (figures \ref{sfig:mmd_entropy_map} and \ref{sfig:wasserstein_entropy_map}). We conclude that Wasserstein- and \gls*{mmd}-\glspl*{gemini} train neural networks not to be overconfident, hence yielding more moderate distributions $\pyx$.

\subsection{Leveraging a manifold geometry}
\label{ssec:exp_moons}
\begin{wrapfigure}{R}{0.5\textwidth}
    \vspace{-5\baselineskip}
    \centering
    \subfloat[$\I^\text{ova}_\text{kl}$ for 2 clusters (MI)]{
        \includegraphics[height=0.1\paperheight,width=0.24\textwidth]{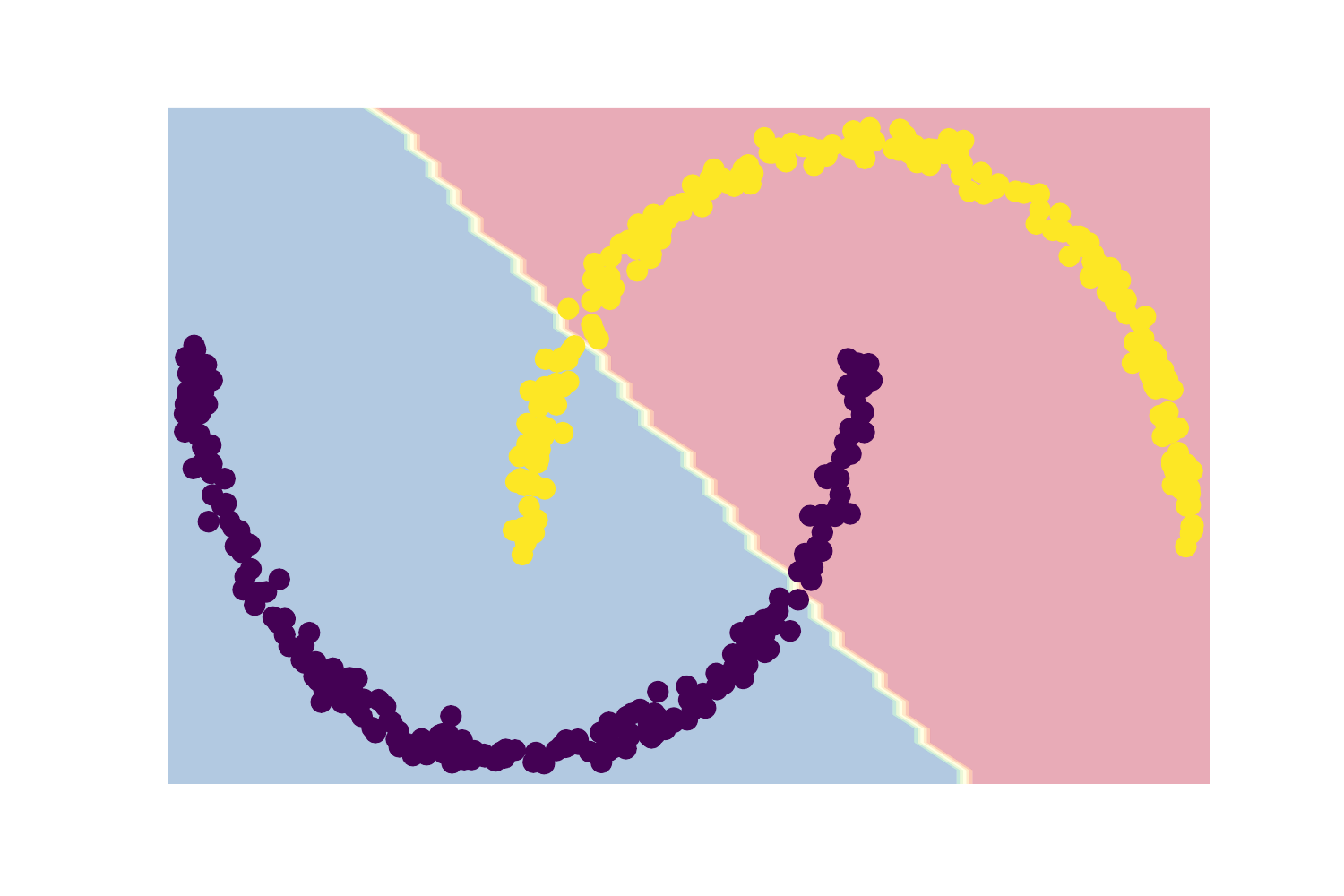}
        \label{sfig:moons_mi_2clusters}
    }
    \subfloat[$\I^\text{ovo}_\mathcal{W}$ for 2 clusters]{
        \includegraphics[height=0.1\paperheight,width=0.24\textwidth]{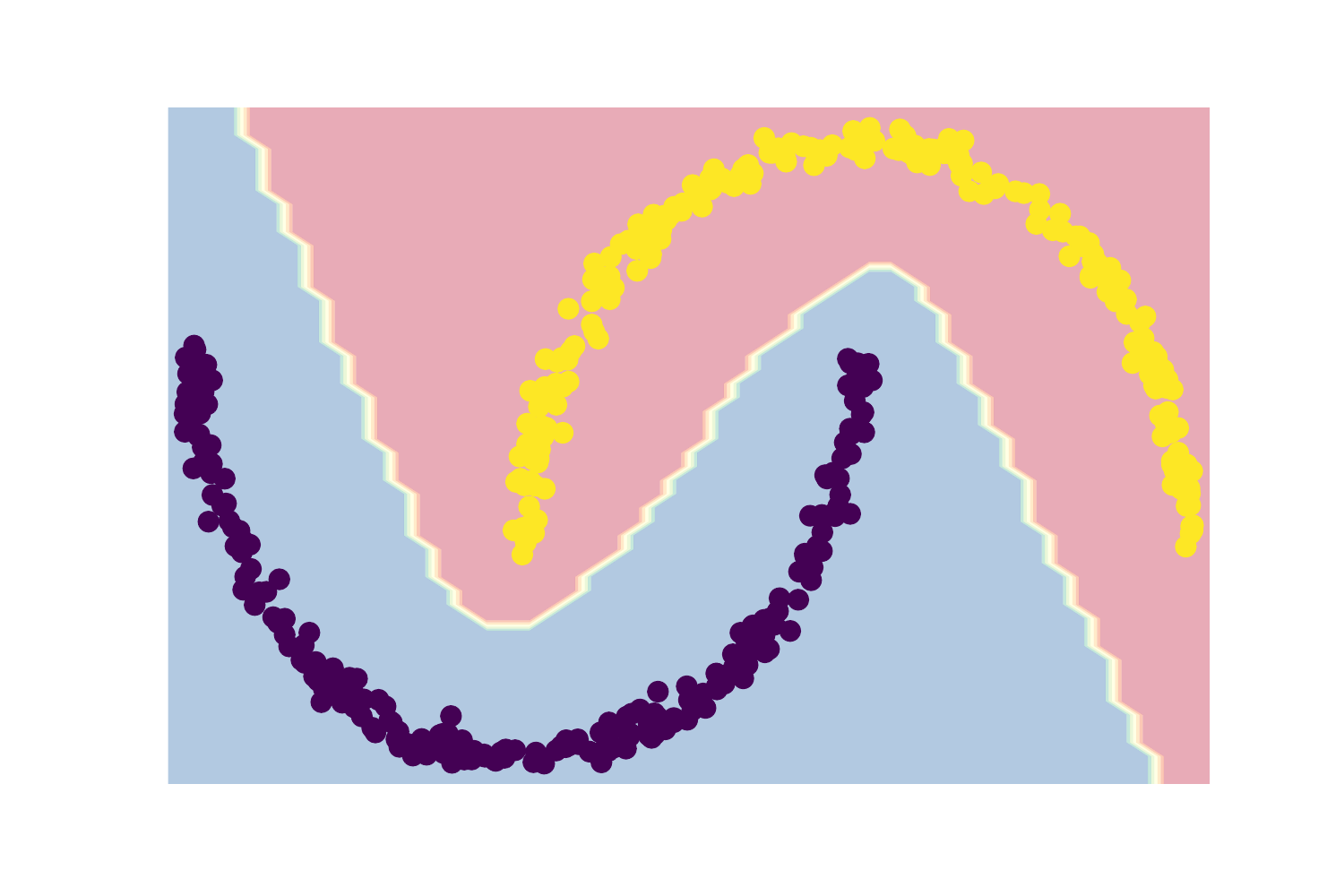}
        \label{sfig:moons_gemini_2clusters}
    }
    \caption{Fitting hand-generated moons using the GEMINI on top of an MLP with 2 clusters to find.}
    \label{fig:moons_decision_boundary}
    \vspace{-2\baselineskip}
\end{wrapfigure}
We highlighted that \gls*{mi} can be maximised without requiring to find the suitable decision boundary. Here, we show how the provided distance to the Wasserstein-\gls*{ovo} \gls*{gemini} can leverage appropriate clustering when we have a good a priori on the data.


\paragraph{The importance of the distance $c$:} We generated a dataset consisting of two facing moons on which we trained a \gls*{mlp} using either the \gls*{mi} or the Wasserstein-\gls*{ovo} \gls*{gemini}. To construct a distance $c$ for the Wasserstein distance, we derived a distance from the Floyd-Warshall algorithm \cite{warshall_theorem_1962,roy_transitivite_1959} on a sparse graph describing neighborhoods of samples. This distance describes how many neighbors are in between two samples, further details are provided in appendix~\ref{app:fw_distance}. We report the different decision boundaries in Figure~\ref{fig:moons_decision_boundary}. We observe that the insight on the neighborhood provided by our distance $c$ helped the \gls*{mlp} to converge to the correct solution with an appropriate decision boundary unlike the \gls*{mi}.  Note that the usual Euclidean distance in the Wasserstein metric would have converged to a solution similar to the \gls*{mi}. Indeed for 2 clusters, the optimal transport plan has a larger value, using a distribution similar to Figure~\ref{sfig:moons_mi_2clusters}, than in Figure~\ref{sfig:moons_gemini_2clusters}. This toy example shows how an insightful metric provided to the Wasserstein distance in \glspl*{gemini} can lead to correct decision boundaries while only designing a discriminative distribution $\pyx$ and a distance $c$.

\begin{wrapfigure}{R}{0.5\textwidth}
    \vspace{-2\baselineskip}
    \centering
    \subfloat[$\I^\text{ova}_\text{kl}$ for 5 clusters (MI)]{
        \includegraphics[height=0.1\paperheight,width=0.24\textwidth]{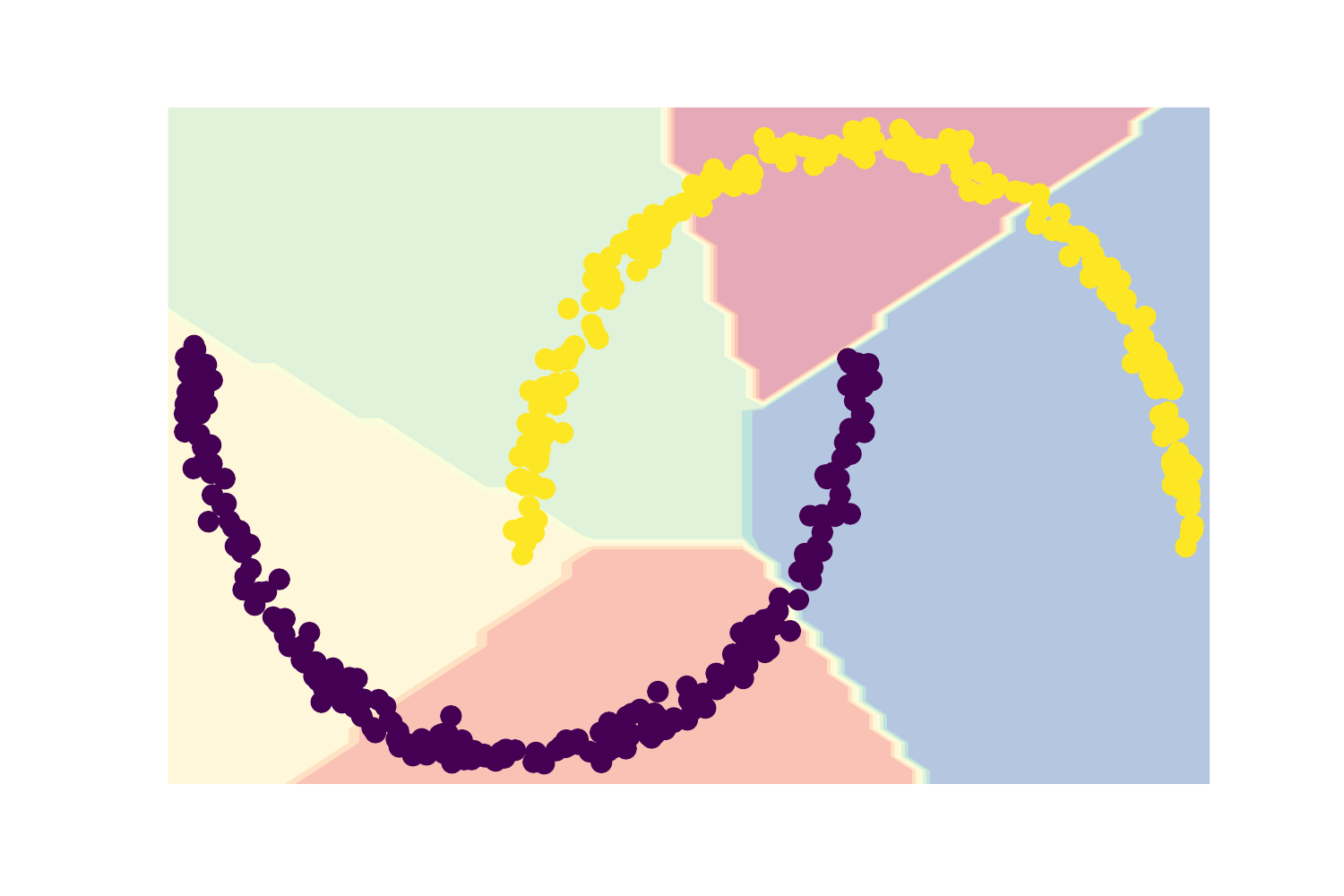}
        \label{sfig:moons_mi_5clusters}
    }
    \subfloat[$\I^\text{ovo}_\mathcal{W}$ for 5 clusters]{
        \includegraphics[height=0.1\paperheight,width=0.24\textwidth]{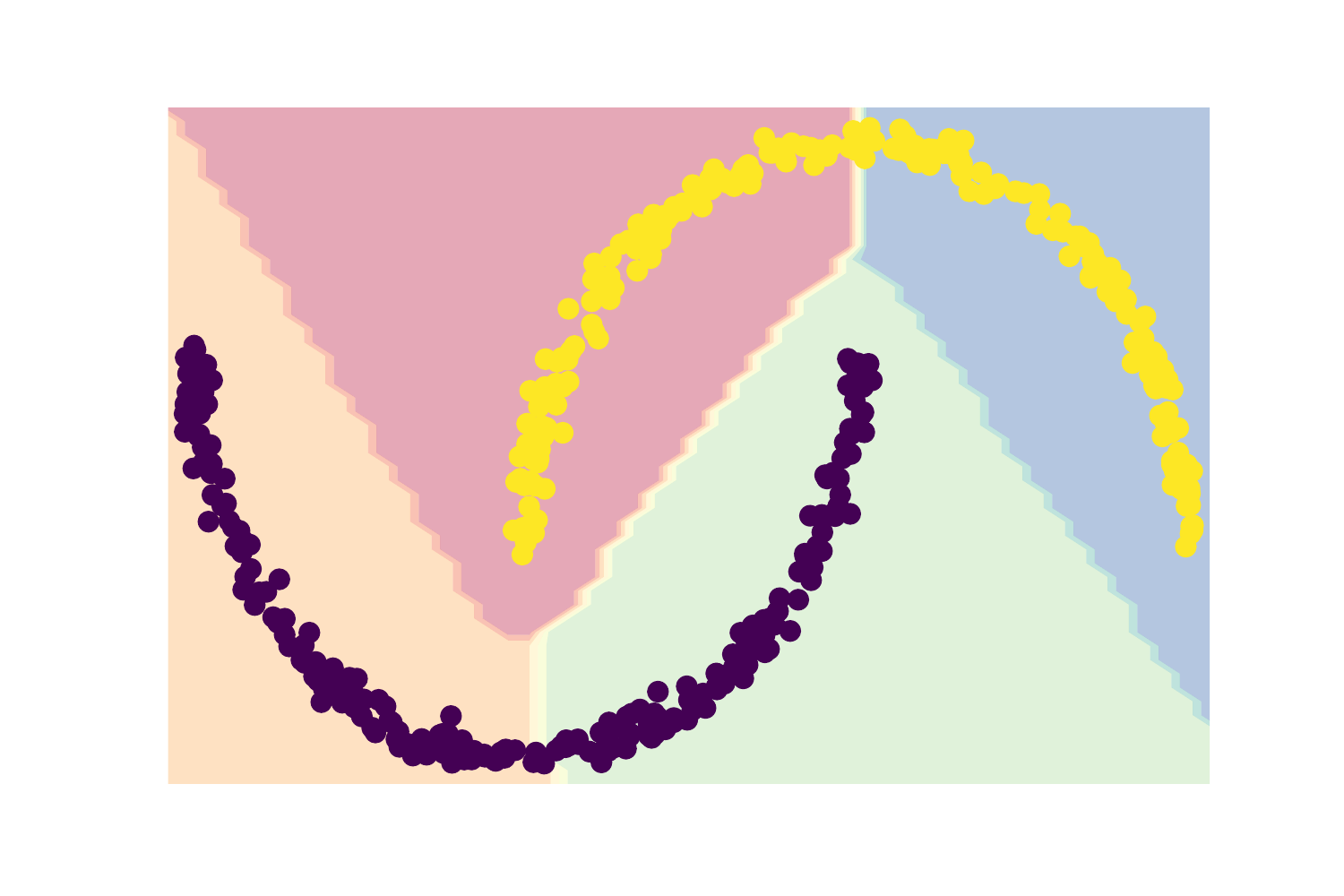}
        \label{sfig:moons_gemini_5clusters}
    }
    \caption{The Wasserstein-ovo model with 5 clusters eventually found 4 unlike the MI that maintained 5 clusters.}
    \label{fig:moons_cluster_emptying}
    \vspace{-2\baselineskip}
\end{wrapfigure}
\paragraph{Not using all clusters} In addition, we highlight an interesting behaviour of all \glspl*{gemini}. During optimisation, it is possible that the model converges to using fewer clusters than the number to find. For example in Figure~\ref{fig:moons_cluster_emptying}, for 5 clusters, the model can converge to 4 balanced clusters and 1 empty cluster (Figure~\ref{sfig:moons_gemini_5clusters}) unlike \gls*{mi} that produced 5 misplaced clusters (Figure~\ref{sfig:moons_mi_5clusters}). Indeed, the entropy on the cluster proportion in the \gls*{mi} forces to use the maximum number of clusters. 
\looseness=-1

\subsection{Fitting MNIST}
\label{ssec:exp_mnist}
\begin{table}[b]
    \caption{ARI for deep neural network trained with GEMINIs on MNIST for 500 epochs. Models were trained either with either 10 clusters to find or 15. We indicate in parentheses the number of used clusters by the model after training.}
    \label{tab:mnist_experiment}
    \centering
    \begin{tabular}{c c c c c c}
    \toprule
        \multicolumn{2}{c}{\multirow{2}[3]{*}{GEMINI}}& \multicolumn{2}{c}{10 clusters}&\multicolumn{2}{c}{15 clusters} \\
        \cmidrule(lr){3-4}\cmidrule(lr){5-6}
        \multicolumn{2}{c}{}&MLP&LeNet-5&MLP&LeNet-5\\
    \midrule
        \multirow{2}{*}{KL} & OvA&0.320 (10)&0.138 (8)&0.271 (15)&0.136 (12)\\
        &OvO&0.348 (7)&0.123 (4)&0.333 (8)&0.104 (4)\\
        
        \multirow{2}{*}{Squared Hellinger} & OvA&0.301 (10)&0.207 (6)&0.224 (13)&0.162 (7)\\
        &OvO&0.287 (10)&0.161 (6)&0.305 (13)&0.157 (7)\\
        
        \multirow{2}{*}{TV} & OvA&0.299 (10)&0.171 (6)&0.277 (15)& 0.140 (6)\\
        &OvO&0.422 (10)&0.161 (9)&0.330 (15)&0.182 (14)\\
        
    \midrule
        
        \multirow{2}{*}{MMD} & OvA&0.373 (10)&0.382 (10)&0.345 (15)&0.381 (15)\\
        &OvO&0.361 (10)&0.373 (10)&0.364 (15)&0.379 (15)\\
        
        \multirow{2}{*}{Wasserstein}& OvA&{\bf 0.471} (10)&{\bf 0.463} (10)&0.390 (15)&{\bf 0.446} (11)\\
        &OvO&0.450 (10)&0.383 (10)&{\bf 0.415} (15)&0.414 (15)\\
        \midrule
        \multicolumn{2}{c}{K-Means}&\multicolumn{2}{c}{0.367}&\multicolumn{2}{c}{0.385}\\
    \bottomrule
    \end{tabular}
\end{table}

We trained a neural network using either \gls*{mi} or \glspl*{gemini}. Following \citet{hu_learning_2017}, we first tried with a \gls*{mlp} with one single hidden layer of dimension 1200. To further illustrate the robustness of the method and its adaptability to other architectures, we also experimented using a LeNet-5 architecture \cite{lecun_gradient_1998} since it is adequate to the MNIST dataset. We report our results in Table~\ref{tab:mnist_experiment}. Since we are dealing with a clustering method, we may not know the number of clusters a priori in a dataset. The only thing that can be said about MNIST is that there are \emph{at least} 10 clusters, one per digit. Indeed, writings of digits could differ leading to more clusters than the number of classes. That is why we further tested the same method with 15 clusters to find in Table~\ref{tab:mnist_experiment}. We first see that the scores of the \gls*{mmd} and Wasserstein \glspl*{gemini} are greater than the \gls*{mi}, with the highest performances for the Wasserstein-\gls*{ova}
We also observe that no $f$-divergence-\gls*{gemini} always yield best ARIs. Nonetheless, we observe better performances in the case of the \gls*{tv} \glspl*{gemini} owing to its bounded gradient. This results in controlled stepsize when doing gradient descent contrarily to \gls*{kl}- and squared Hellinger-\glspl*{gemini}. Notice that the change of architecture from a \gls*{mlp} to a LeNet-5 unexpectedly halves the scores for the $f$-divergences. We believe this drop is due to the change of notion of neighborhood implied by the network architecture.

\subsection{Cifar10 clustering using a SIMCLR-derived kernel}
\label{ssec:exp_cifar10}
\begin{table}[b]
    \caption{ARI score of models trained for 200 epochs on CIFAR10 with different architectures using GEMINIs. The kernel for the MMD is either a linear kernel or the dot product between features extracted from a pretrained SIMCLR model. Both the Euclidean norm between images and SIMCLR features are considered for the Wasserstein metric. We report the ARI of related works when not using data augmentation for comparison.*: scores reported from Li et al., (2021)}
    \label{tab:simclr_kernel}
    \centering
    \begin{tabular}{c c c c c c c c c c}
    \toprule
    \multirow{2}{*}{Architecture}&No kernel&\multicolumn{4}{c}{Linear kernel / $\ell_2$ norm}&\multicolumn{4}{c}{SIMCLR \cite{chen_simple_2020}}\\
    \cmidrule(l){2-2}\cmidrule(lr){3-6}\cmidrule{7-10}
    &$\I_\text{KL}^\text{ova}$&$\I_\text{MMD}^\text{ova}$&$\I_\text{MMD}^\text{ovo}$&$\I_\mathcal{W}^\text{ova}$&$\I_\mathcal{W}^\text{ovo}$&$\I_\text{MMD}^\text{ova}$&$\I_\text{MMD}^\text{ovo}$&$\I_\mathcal{W}^\text{ova}$&$\I_\mathcal{W}^\text{ovo}$\\
    \midrule
    LeNet-5&0.026&0.049&0.048&0.043&0.041&{\bf 0.157}&0.145&0.079&0.138\\
    Resnet-18&0.008&0.047&0.044&0.037&0.036&0.122&{\bf 0.145}&0.052&0.080\\
    \midrule
    \multicolumn{3}{c}{KMeans (images / SIMCLR)}&0.041&0.147&\multicolumn{3}{c}{CC \cite{li_contrastive_2021}}&\multicolumn{2}{c}{0.030}\\
    \multicolumn{3}{c}{IDFD \cite{tao_clustering-friendly_2021}}&\multicolumn{2}{c}{0.060}&\multicolumn{3}{c}{JULE \cite{yang_joint_2016}*}&\multicolumn{2}{c}{0.138}\\
    \bottomrule
    \end{tabular}
\end{table}

To further illustrate the benefits of the kernel or distance provided to \glspl*{gemini}, we continue the same experiment as in section~\ref{ssec:exp_mnist}. However, we focus this time on the CIFAR10 dataset. As improved distance, we chose a linear kernel and $\ell_2$ norm between features extracted from a pretrained SIMCLR model \cite{chen_simple_2020}. We provide results for two different architectures: LeNet-5 and ResNet-18 both trained from scratch on raw images, the latter being a common choice of models in deep clustering literature \cite{van_gansbeke_scan_2020,tao_clustering-friendly_2021}. We report the results in Table~\ref{tab:simclr_kernel} and provide the baseline of \gls*{mi}. We also write the baselines from related works when not using data augmentations to make a fair comparison. Indeed, models trained with \glspl*{gemini} do not use data augmentation: only the architecture and the kernel or distance function in the data space plays a role. We observe here that the choice of kernel or distance can be critical in \glspl*{gemini}. Indeed, while the Euclidean norm between images does not provide insights on how images of cats and dogs are far as shown by K-Means, features derived from SIMCLR carry much more insight on the proximity of images. This shows that the performances of \glspl*{gemini} depend on the quality of distance functions. Interestingly, we observe that for the Resnet-18 using SIMCLR features to guide \glspl*{gemini} was not as successful as it has been on the LeNet-5. We believe that the ability of this network to draw any decision boundary makes it equivalent to a categorical distribution model as in Sec.~\ref{ssec:exp_categorical}. Finally, to the best of our knowledge, we are the first to train from scratch a standard discriminative neural network on CIFAR raw images without using labels or direct data augmentations, while getting sensible clustering results. However, other recent methods achieve best scores using data augmentations which we do not \cite{park_improving_2021}.

\section{Conclusion}
\label{sec:conclusions}
We highlighted that the choice of distance at the core of \gls*{mi} can alter the performances of deep learning models when used as an objective for a deep discriminative clustering. We first showed that \gls*{mi} maximisation does not necessarily reflect the best decision boundary in clustering. We introduced the \gls*{gemini}, a method which only needs the specification of a neural network and a kernel or distance in the data space. Moreover, we showed how the notion of neighborhood built by the neural network can affect the clustering, especially for \gls*{mi}. To the best of our knowledge, this is the first method that trains single-stage neural networks from scratch using neither data augmentations nor regularisations, yet achieving good clustering performances. We emphasised that \glspl*{gemini} are only searching for a maximum number of clusters: after convergence some may be empty. However, we do not have insights to explain this convergence which is part of future work. Finally, we introduced several versions of \glspl*{gemini} and would encourage the \gls*{mmd}-\gls*{ova} or Wasserstein-\gls*{ova} as a default choice, since it proves to both incorporate knowledge from the data using a kernel or distance while remaining the less complex than \gls*{mmd}-\gls*{ovo} and Wasserstein-\gls*{ovo} in time and memory. \gls*{ovo} versions could be privileged for fine-tuning steps. Future works could include an optimisation of the time performances of the Wasserstein-\gls*{ovo} to make it more competitive.

\section*{Acknowledgements}
This work has been supported by the French government, through the 3IA C\^ote d'Azur, Investment in the Future, project managed by the National Research Agency (ANR) with the reference number ANR-19-P3IA-0002. We would also like to thank the France Canada Research Fund (FFCR) for their contribution to the project. This work was partly supported by EU Horizon 2020 project AI4Media, under contract no. 951911. The authors are grateful to the OPAL infrastructure from Université Côte d'Azur for providing resources and support.

\bibliography{bib}

\newpage
\section*{Checklist}

\begin{enumerate}

\item For all authors...
\begin{enumerate}
  \item Do the main claims made in the abstract and introduction accurately reflect the paper's contributions and scope?
    \answerYes{We clearly state that we focus on mutual information for clustering and how we improve it.}
  \item Did you describe the limitations of your work?
    \answerYes{We describe shortcomings in Sec.~\ref{ssec:exp_cifar10},\ref{sec:conclusions} and provide more in App.~\ref{app:exp_complexity}.}
  \item Did you discuss any potential negative societal impacts of your work?
    \answerNA{}
  \item Have you read the ethics review guidelines and ensured that your paper conforms to them?
    \answerYes{Yes, we confirm having read the ethics review guidelines and conform to it.}
\end{enumerate}

\item If you are including theoretical results...
\begin{enumerate}
  \item Did you state the full set of assumptions of all theoretical results?
    \answerYes{We clearly highlighted that we compute the GEMINI between a continuous variable, namely the data, and a discrete variable, namely the cluster assignment. The framework was detailed precisely in Sec.~\ref{ssec:discriminative_clustering} and all distances have been introduced.}
        \item Did you include complete proofs of all theoretical results?
    \answerYes{Please refer to appendices~\ref{app:mi_convergence}, \ref{app:wasserstein_convergence} or \ref{app:deriving_geminis} for proofs of our claims.}
\end{enumerate}

\item If you ran experiments...
\begin{enumerate}
  \item Did you include the code, data, and instructions needed to reproduce the main experimental results (either in the supplemental material or as a URL)?
    \answerYes{Link to github will be available upon release of paper and a sample is zipped aside in the supplementary material.}
  \item Did you specify all the training details (e.g., data splits, hyperparameters, how they were chosen)?
    \answerYes{Data generation processes are reported in supplementary materials and main parameters are described at the head of Sec.~\ref{sec:experiments}.}
        \item Did you report error bars (e.g., with respect to the random seed after running experiments multiple times)?
    \answerYes{When applicable due to multiple runs in table form, e.g. Table~\ref{tab:mnist_experiment}.}
        \item Did you include the total amount of compute and the type of resources used (e.g., type of GPUs, internal cluster, or cloud provider)?
    \answerYes{We mention it at the beginning of Sec.~\ref{sec:experiments}}
\end{enumerate}

\item If you are using existing assets (e.g., code, data, models) or curating/releasing new assets...
\begin{enumerate}
  \item If your work uses existing assets, did you cite the creators?
    \answerYes{All packages for the experiments are mentionned in Appendix~\ref{app:requirements}}
  \item Did you mention the license of the assets?
    \answerYes{We are under licenses of MIT (NumPy, POT,), BSD (torch,pandas, scikit-learn) and the python software foundation (matplotlib, python).}
  \item Did you include any new assets either in the supplemental material or as a URL?
    \answerYes{The code once released will contain the specific file \verb+losses.py+ that will contain only the definitions of GEMINIs, alongside all configurations to reproduce the experiments.}
  \item Did you discuss whether and how consent was obtained from people whose data you're using/curating?
    \answerNA{}
  \item Did you discuss whether the data you are using/curating contains personally identifiable information or offensive content?
    \answerNA{}
\end{enumerate}

\item If you used crowdsourcing or conducted research with human subjects...
\begin{enumerate}
  \item Did you include the full text of instructions given to participants and screenshots, if applicable?
    \answerNA{}
  \item Did you describe any potential participant risks, with links to Institutional Review Board (IRB) approvals, if applicable?
    \answerNA{}
  \item Did you include the estimated hourly wage paid to participants and the total amount spent on participant compensation?
    \answerNA{}
\end{enumerate}

\end{enumerate}

\newpage
\appendix

\section{Demonstration of the convergence to 0 of the MI for a Gaussian Mixture}
\label{app:mi_convergence}
\newcommand{\nmuA}{\mathcal{N}(x|\mu_0,\sigma^2)}
\newcommand{\nmuB}{\mathcal{N}(x|\mu_1,\sigma^2)}

\label{app:mi_boundaries}
\subsection{Models definition}
Let us consider a mixture of two Gaussian distributions, both with different means $\mu_0$ and $\mu_1$, s.t. $\mu_0<\mu_1$ and of same standard deviation $\sigma$:

\begin{equation*}
p(x|y=0) = \nmuA, p(x|y=1) = \nmuB ,
\end{equation*}

where $y$ is the cluster assignment. We take balanced clusters proportions, i.e. $p(y=0)=p(y=1)=\frac{1}{2}$. This first model is the basis that generated the complete dataset $p(x)$. When performing clustering with our discriminative model, we are not aware of the distribution. Consequently: we create other models. We want to compute the difference of mutual information between two decision boundaries that discriminative models $p_\theta(y|x)$ may yield.

We define two decision boundaries: one which splits evenly the data space called $p_A$ and another which splits it on a closed set $p_B$:

\begin{equation*}\label{eq:p_a}
p_A(y=1|x) = \left\{ \begin{array}{c c}
1-\epsilon&x>\frac{\mu_1-\mu_0}{2}\\
\epsilon&\text{otherwise}
\end{array}\right. ,
\end{equation*}

\begin{equation}\label{eq:p_b}
p_B(y=1|x) = \left\{ \begin{array}{c c}
1-\epsilon&x \in [\mu_0, \mu_1]\\
\epsilon&\text{otherwise}
\end{array}\right. .
\end{equation}

Our goal is to show that both models $p_A$ and $p_B$ will converge to the same value of mutual information as $\epsilon$ converges to 0.

\subsection{Computing cluster proportions}
\subsubsection{Cluster proportion of the correct decision boundary}

To compute the cluster proportions, we estimate with samples $x$ from the distribution $p_\text{data}(x)$. Since we are aware for this demonstration of the true nature of the data distribution, we can use $p(x)$ for sampling. Consequently, we can compute the two marginals:

\begin{align*}
p_A(y=1) &= \int_\mathcal{X} p(x) p_A(y=1|x)dx ,\\
&= \int_{-\infty}^{\frac{\mu_1-\mu_0}{2}}p(x)\epsilon dx + \int_{\frac{\mu_1-\mu_0}{2}}^{+\infty}p(x)(1-\epsilon)dx ,\\
&=\epsilon \left(\int_{-\infty}^{\frac{\mu_1-\mu_0}{2}}p(x)dx \right) + (1-\epsilon) \left( \int_{\frac{\mu_1-\mu_0}{2}}^{+\infty}p(x)dx\right) ,\\
&= \frac{1}{2} .
\end{align*}

\subsubsection{Cluster proportion of the misplaced decision boundary}
For the misplaced decision boundary, the marginal is different:

\begin{equation}\begin{split}
p_B(y=1) &= \int_\mathcal{X} p(x) p_B(y=1|x)dx ,\\
&= \epsilon\left(\int_{-\infty}^{\mu_0}p(x)dx + \int_{\mu_1}^{+\infty}p(x)dx\right) + (1-\epsilon) \int_{\mu_0}^{\mu_1}p(x)dx ,\\
&=\epsilon \left(1-\int_{\mu_0}^{\mu_1}p(x)dx\right) + (1-\epsilon)\int_{\mu_0}^{\mu_1}p(x)dx .
\end{split}\end{equation}

Here, we simply introduce a new variable named $\beta$ that will be a shortcut for noting the proportion of data between $\mu_0$ and $\mu_1$:

\begin{equation*}
\beta = \int_{\mu_0}^{\mu_1}p(x)dx .
\end{equation*}

And so can we simply write the cluster proportion of decision boundary model B as:

\begin{equation*}\label{eq:py_b}
p_B(y=1) = \epsilon (1-\beta) + (1-\epsilon)\beta ,
\end{equation*}

Leading to the summary of proportions in Table~\ref{tab:proportions}. For convenience, we will write the proportions of model B using the shortcuts:

\begin{equation*}\label{eq:pi_b}
\pi_B = p_B(y=1) = \epsilon + \beta(1-2\epsilon) ,
\end{equation*}
\begin{equation*}
\bar{\pi}_B = p_B(y=0) = 1-\epsilon - \beta(1-2\epsilon) .
\end{equation*}

\begin{table}[hbt]
\caption{Proportions of clusters for models A and B}\label{tab:proportions}
\centering
\begin{tabular}{c c c}
\toprule
$\mathcal{M}$&A&B\\
\cmidrule(lr){2-3}
$p_\mathcal{M}(y=1)$&$\frac{1}{2}$&$\epsilon+\beta(1-2\epsilon)$\\
$p_\mathcal{M}(y=0)$&$\frac{1}{2}$&$1-\epsilon-\beta(1-2\epsilon)$\\
\bottomrule
\end{tabular}
\end{table}

\subsection{Computing the KL divergences}
\subsubsection{Correct decision boundary}

We first start by computing the Kullback-Leibler divergence for some arbitrary value of $x\in\mathbb{R}$:

\begin{equation*}
D_\text{KL}(p_A(y|x)||p_A(y))= \sum_{i=0}^1 p_A(y=i|x) \log{\frac{p_A(y=i|x)}{p_A(y=i)}} .
\end{equation*}

We now need to detail the specific cases, for the value of $p(y=i|x)$ is dependent on $x$. We start $\forall x <\frac{\mu_1-\mu_0}{2}$:

\begin{align*}
D_\text{KL}(p_A(y|x)||p_A(y))&= p_A(y=0|x)\log{\frac{p_A(y=0|x)}{\frac{1}{2}}} + p_A(y=1|x)\log{\frac{p_A(y=1|x)}{\frac{1}{2}}} ,\\
&=(1-\epsilon)\log{2(1-\epsilon)} + \epsilon \log{2\epsilon} .
\end{align*}

The opposite case, $\forall x\geq \frac{\mu_1-\mu_0}{2}$ yields:
\begin{align*}
D_\text{KL}(p_A(y|x)||p_A(y))&= p_A(y=0|x)\log{\frac{p_A(y=0|x)}{\frac{1}{2}}} + p_A(y=1|x)\log{\frac{p_A(y=1|x)}{\frac{1}{2}}} ,\\
&=\epsilon\log{2\epsilon} + (1-\epsilon) \log{2(1-\epsilon)} .
\end{align*}

Since both cases are equal, we can write down:

\begin{equation}\label{eq:correct_kl_div}
D_\text{KL}(p_A(y|x)||p_A(y)) = \epsilon\log{2\epsilon} + (1-\epsilon)\log{2(1-\epsilon)} ,\forall x\in\mathbb{R} .
\end{equation}

\subsubsection{Misplaced boundary}

We proceed to the same detailing of the Kullback-Leibler divergence computation for the misplaced decision boundary. We start with the set $x\in [\mu_0,\mu_1]$:

\begin{align*}
D_\text{KL}(p_B(y|x)||p_B(y))&= p_B(y=0|x) \log{\frac{p_B(y=0|x)}{p_B(y=0)}} + p_B(y=1|x)\log{\frac{p_B(y=1|x)}{p_B(y=1)}} ,\\
&=\epsilon \log{\frac{\epsilon}{\bar{\pi}_B}} + (1-\epsilon) \log{\frac{1-\epsilon}{\pi_B}} .
\end{align*}

When $x$ is out of this set, the divergence becomes:

\begin{align*}
D_\text{KL}(p_B(y|x)||p_B(y))&=p_B(y=0|x) \log{\frac{p_B(y=0|x)}{p_B(y=0)}} + p_B(y=1|x)\log{\frac{p_B(y=1|x)}{p_B(y=1)}} ,\\
&=(1-\epsilon)\log{\frac{1-\epsilon}{\bar{\pi}_B}} + \epsilon \log{\frac{\epsilon}{\pi_B}} .
\end{align*}

To fuse the two results, we will write the KL divergence as such:

\begin{equation}\label{eq:odd_kl_div}
D_\text{KL}(p_B(y|x)||p_B(y)) = \epsilon\log{\epsilon}+ (1-\epsilon)\log(1-\epsilon) - C(x) ,\forall x\in\mathbb{R} ,
\end{equation}

where $C(x)$ is a constant term depending on $x$ defined by:

\begin{equation}\label{eq:odd_constant}
C(x)=\left\{\begin{array}{c c}
\epsilon\log{\bar{\pi}_B} + (1-\epsilon)\log{\pi_B}&x\in[\mu_0,\mu_1]\\
\epsilon\log{\pi_B} + (1-\epsilon)\log{\bar{\pi}_B}&x\in\mathbb{R}\setminus[\mu_0,\mu_1]\\
\end{array}  .\right.
\end{equation}

For simplicity of later writings, we will shorten the notations by:

\begin{equation*}
C(x)=\left\{ \begin{array}{c c}
\alpha_1&x\in[\mu_0,\mu_1]\\
\alpha_0&x\in\mathbb{R}\setminus[\mu_0,\mu_1]
\end{array}\right. .
\end{equation*}

\subsection{Evaluating the mutual information}

\subsubsection{Correct decision boundary}

We inject the value of the Kullback-Leibler divergence from Eq.~(\ref{eq:correct_kl_div}) inside an expectation performed over the data distribution $p_\text{data}(x)$:

\begin{align}
\mathcal{I}_A(X;Y) &= \mathbb{E}_{x\sim p_\text{data}(x)}\left[ D_\text{KL}(p_A(y|x)||p_A(y))\right] ,\\
&=\int_\mathcal{X}p(x) \left(\epsilon \log(2\epsilon) + (1-\epsilon)\log(2(1-\epsilon)) \right)dx ,\\
&=\epsilon \log(2\epsilon) + (1-\epsilon)\log(2(1-\epsilon)) .\label{eq:mi_a}
\end{align}

Since the KL divergence was independent of $x$, we could leave the constant outside of the integral which is equal to 1.

We can assess the coherence of Eq.~(\ref{eq:mi_a}) since its limit as $\epsilon$ approaches 0 is $\log 2$. In terms of bits, this is the same as saying that the information on $X$ directly gives us information on the $Y$ of the cluster.

\subsubsection{Odd decision boundary}

We inject the value of the KL divergence from Eq.~(\ref{eq:odd_kl_div}) inside the expectation of the mutual information:

\begin{align*}
\mathcal{I}_B(X;Y) &= \mathbb{E}_{x\sim p_\text{data}(x)}\left[ D_\text{KL}(p_B(y|x)||p_B(y))\right] ,\\
&=\int_{\mathcal{X}}p(x)\left(\epsilon\log\epsilon +(1-\epsilon)\log(1-\epsilon) -C(x)\right) ,dx\\
&=\epsilon\log\epsilon +(1-\epsilon)\log(1-\epsilon) - \int_{\mathcal{X}}p(x)C(x)dx .
\end{align*}

The first terms are constant with respect to $x$ and the integral of $p(x)$ over $\mathcal{X}$ adds up to 1. We finally need to detail the expectation of the constant $C(x)$ from Eq.~(\ref{eq:odd_constant}):

\begin{align*}
\mathbb{E}_x[C(x)] &= \int_{-\infty}^{\mu_0}C(x)p(x)dx + \int_{\mu_0}^{\mu_1}C(x)p(x)dx + \int_{\mu_1}^{+\infty}C(x)p(x)dx ,\\
&= \alpha_0\left(\int_{-\infty}^{\mu_0}p(x)dx + \int_{\mu_1}^{+\infty}p(x)dx \right) + \alpha_1\int_{\mu_0}^{\mu_1}p(x)dx ,\\
&=\alpha_0(1-\beta) + \alpha_1\beta .
\end{align*}

This can be further improved by unfolding the description of $\alpha_0$ and $\alpha_1$ from Eq.~(\ref{eq:odd_constant}):

\begin{align*}
\alpha_0(1-\beta)+\beta\alpha_1&= \alpha_0 +\beta(\alpha_1-\alpha_0),\\
&=\epsilon\log{\pi_B}+(1-\epsilon)\log{\bar{\pi}_B} + \beta\left[\epsilon\log{\bar{\pi}_B}+(1-\epsilon)\log{\pi_B}\right.\\&\quad\left.-\epsilon\log{\pi_B}- (1-\epsilon)\log{\bar{\pi}_B}\right],\\
&= \left[1-\epsilon+\beta\epsilon-\beta+\beta\epsilon\right]\log{\bar{\pi}_B} + \left[\epsilon+\beta-\beta\epsilon-\beta\epsilon\right]\log{\pi_B},\\
&=\log{\bar{\pi}_B} + \left[2\beta\epsilon-\beta-\epsilon\right]\log{\frac{\bar{\pi}_B}{\pi_B}} .
\end{align*}

We can finally write down the mutual information for the model B:

\begin{equation}\label{eq:mi_b}
\mathcal{I}_B(X;Y) = \epsilon\log{\epsilon} +(1-\epsilon)\log(1-\epsilon) -\log{\bar{\pi}_B} - \left[2\beta\epsilon-\beta-\epsilon\right]\log{\frac{\bar{\pi}_B}{\pi_B}} .
\end{equation}

\subsection{Differences of mutual information}

Now that we have the exact value of both mutual informations, we can compute their differences:

\begin{align*}
\Delta_\mathcal{I} &= \mathcal{I}_A(X;Y) - \mathcal{I}_B(X;Y),\\
&= \epsilon\log(2\epsilon) + (1-\epsilon)\log(2(1-\epsilon)) - \epsilon\log{\epsilon} -(1-\epsilon)\log(1-\epsilon)\\&\quad+\log{\bar{\pi}_B} + \left[2\beta\epsilon-\beta-\epsilon\right]\log{\frac{\bar{\pi}_B}{\pi_B}} ,\\
&=\epsilon\log{2}+(1-\epsilon)\log2 +\log{\bar{\pi}_B} + \left[2\beta\epsilon-\beta-\epsilon\right]\log{\frac{\bar{\pi}_B}{\pi_B}} .
\end{align*}

We then deduce how the difference of mutual information evolves as the decision boundary becomes sharper, i.e. when $\epsilon$ approaches 0:

\begin{equation*}
\lim_{\epsilon\rightarrow 0}\Delta_\mathcal{I} = \log{2} + \log{\bar{\pi}_B} -\beta\log{\frac{\bar{\pi}_B}{\pi_B}} .
\end{equation*}

However, the cluster proportions by B $\pi_B$ also take a different value as $\epsilon$ approaches 0. Recalling Eq.~(\ref{eq:p_b}):

\begin{equation*}
\lim_{\epsilon\rightarrow 0} p_B(y=1) = \beta , \lim_{\epsilon\rightarrow 0} p_B(y=0) = 1-\beta .
\end{equation*}

And finally can we write that:

\begin{equation}\begin{split}
\lim_{\epsilon\rightarrow 0} \Delta_\mathcal{I} &= \log{2} + \log(1-\beta) - \beta \log\frac{1-\beta}{\beta} ,\\
&=\log{2} + (1-\beta)\log(1-\beta) + \beta \log{\beta} ,\\
&=\log{2} -\mathcal{H}(\beta) .
\end{split}\end{equation}

\begin{figure}[hbt]
    \centering
    \subfloat[Differences of MI between models A and B]{
        \includegraphics[width=0.45\linewidth]{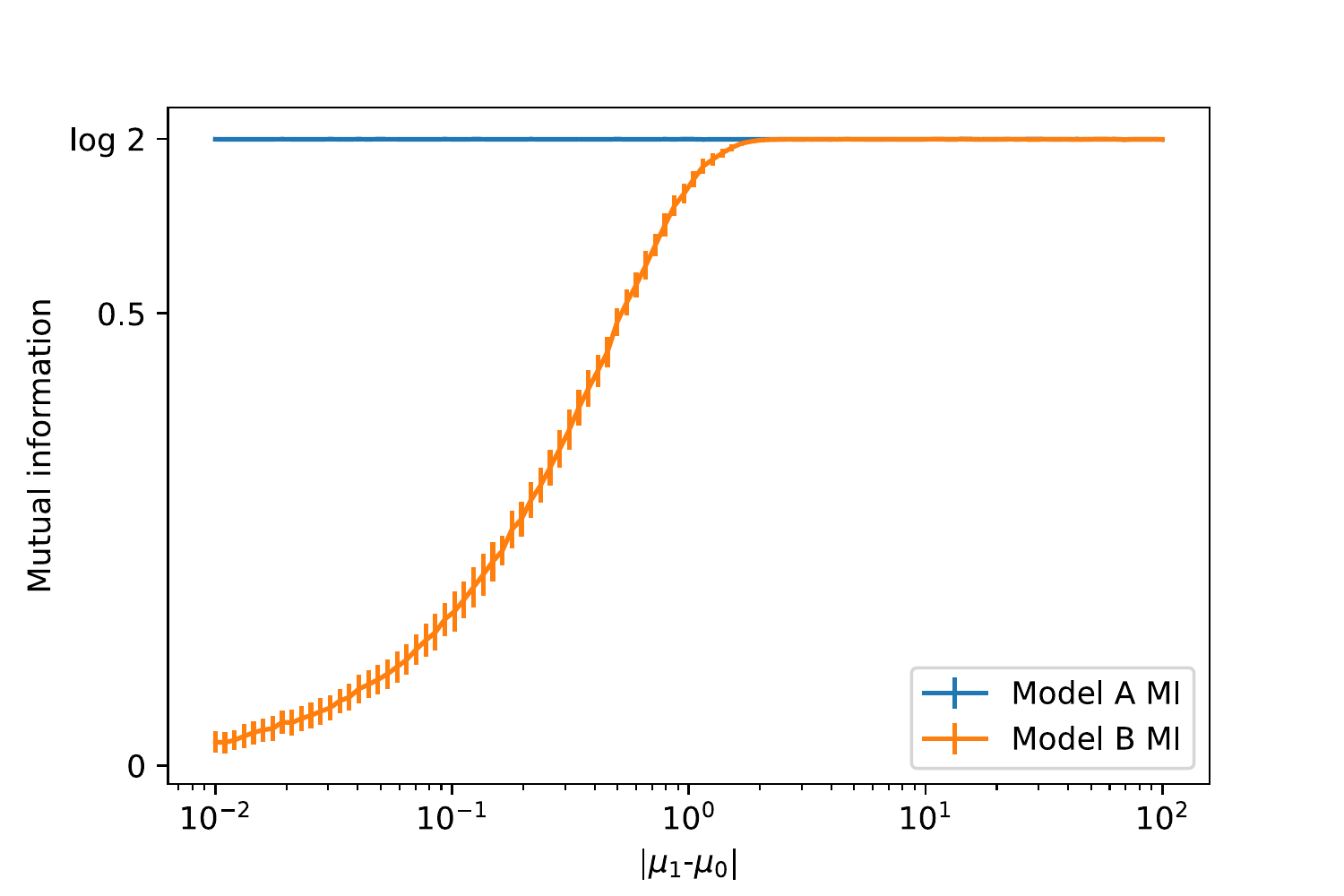}
        \label{sfig:mi_convergence_differences}
    }
    \subfloat[Gaussian mixture distribution $p(x)$ with proportion $\beta$ in between the two means $\mu_0$ and $\mu_1$]{
        \includegraphics[width=0.45\linewidth]{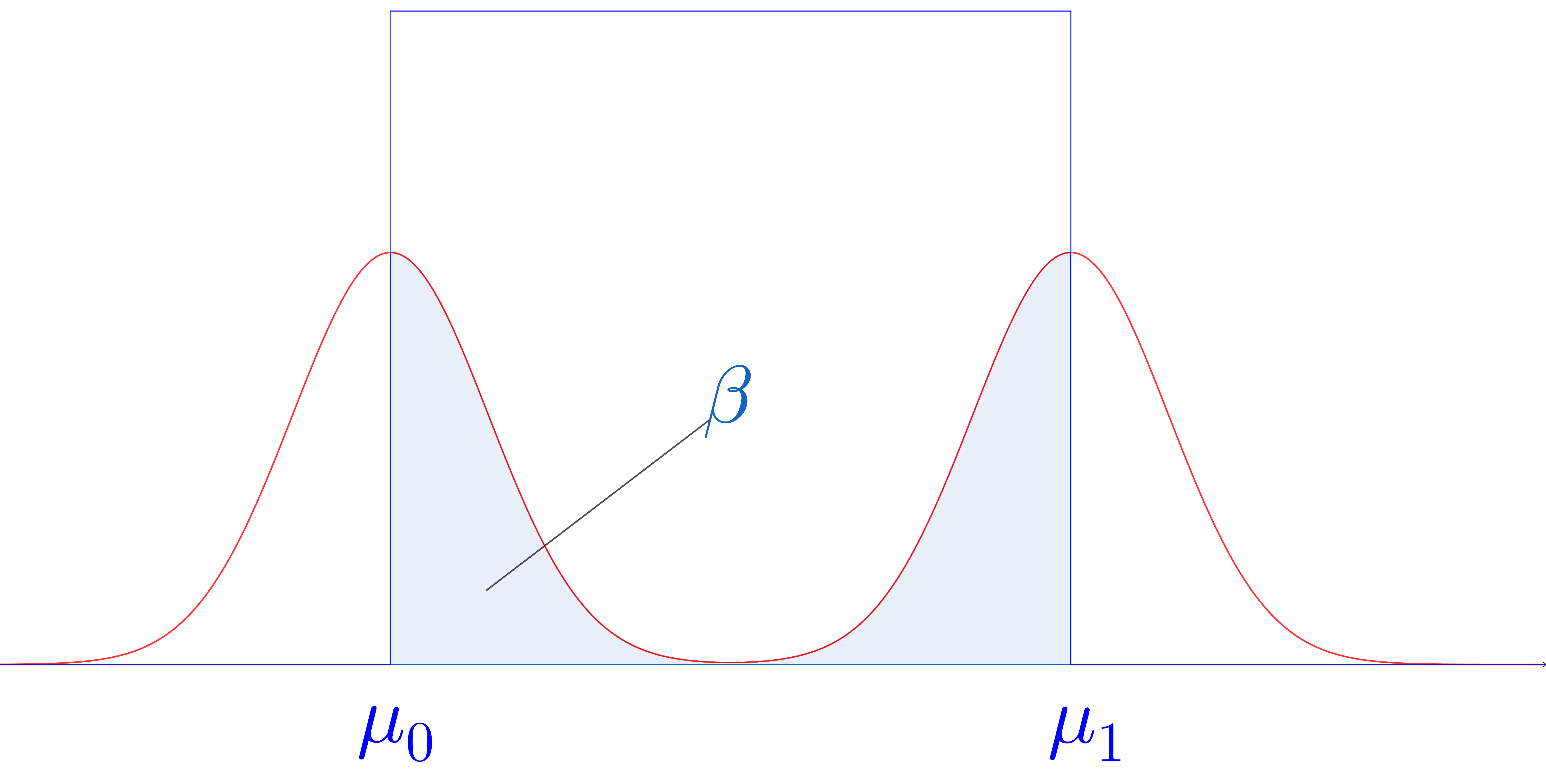}
        \label{sfig:recap_gaussian_mixture}
    }
    \caption{Value of the mutual information for the two models splitting a Gaussian mixture dependening on the distance between the two means $\mu_0$ and $\mu_1$ of the two generating Gaussian distributions. We estimate the MI by computing it 50 times on 1000 samples drawn from the Gaussian mixture.}
    \label{fig:mi_convergence_differences}
\end{figure}

To conclude, as the decision boundaries turn sharper, i.e. when $\epsilon$ approaches 0, the difference of mutual information between the two models is controlled by the entropy of proportion of data $\beta$ between the two means $\mu_0$ and $\mu_1$. We know that for binary entropies, the optimum is reached for $\beta=0.5$. In other words having $\mu_0$ and $\mu_1$ distant enough to ensure balance of proportions between the two clusters of model B leads to a difference of mutual information equal to 0. We experimentally highlight this convergence in Figure~\ref{fig:mi_convergence_differences} where we compute the mutual information of models A and B as the distance between the two means $\mu_0$ and $\mu_1$ increases in the Gaussian distribution mixture.

\section{A geometrical perspective on OvA and OvO GEMINIs}
\label{app:ovo_and_ova}
\begin{figure}
\centering
\includegraphics[width=0.5\linewidth]{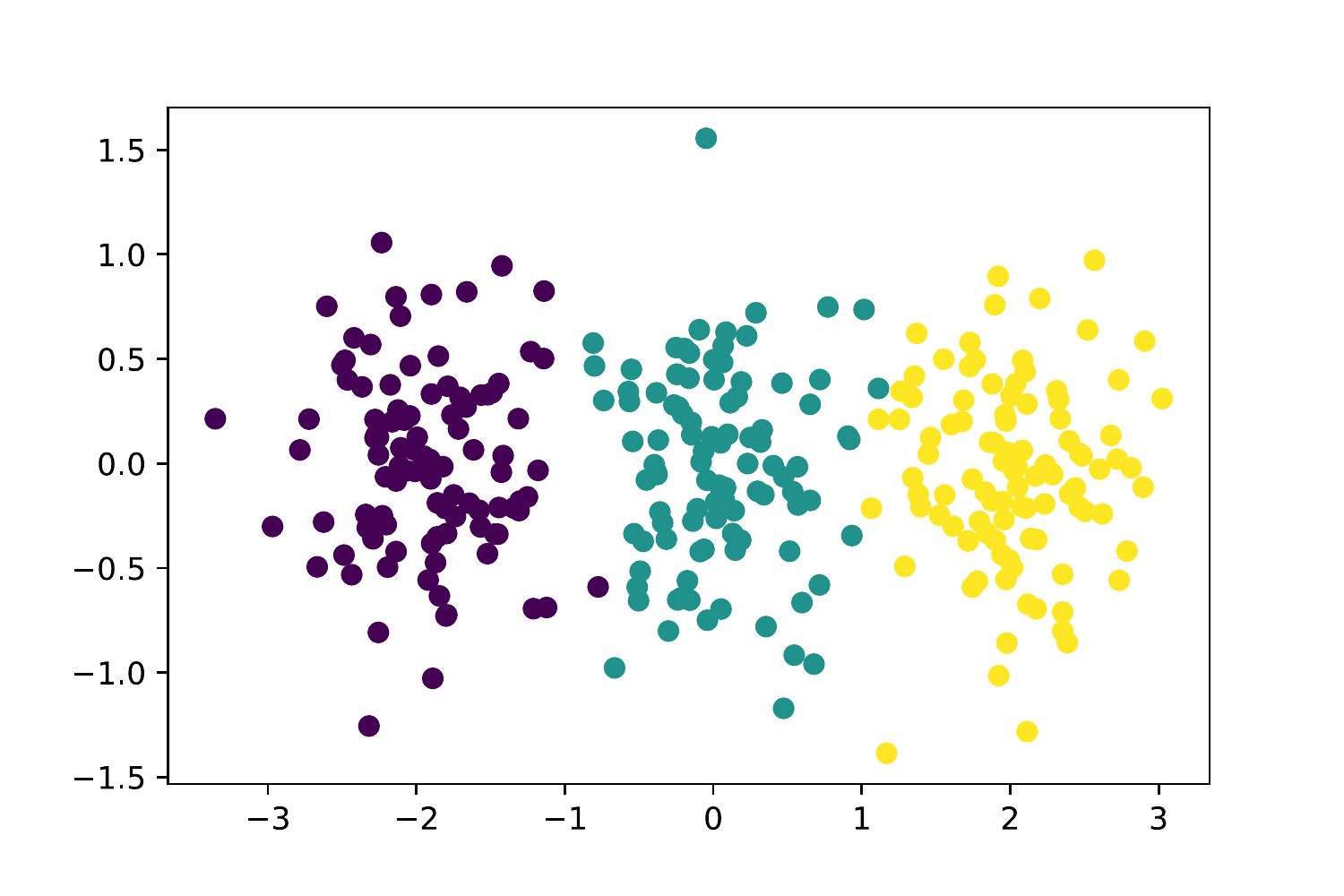}
\caption{Here, 3 clusters of equal proportions from isotropic Gaussian distributions are located in -2, 0 and 2 on the x-axis, with small covariance. The complete data distribution hence has its expectation in 0 on the x-axis. Consequently, maximising the MMD OvA GEMINI can lead to 2 clusters at best while the MMD OvO is able to see all 3 clusters.}\label{fig:example_ova_ovo}
\end{figure}

Considering the topology of the data space through a kernel in the case of the MMD or a distance in the case of the Wasserstein metric implies that we can effectively measure how two distributions are close to another. In the formal design of the mutual information, the distribution of each cluster $p(\vec{x}|y)$ is compared to the complete data distribution $p(\vec{x})$. Therefore, if one distribution of a specific cluster $p(\vec{x}|y)$ were to look alike the data distribution $p(\vec{x})$, for example up to a constant in some areas of the space, then its distance to the data distribution could be 0, making it unnoticed when maximising the OvA GEMINI.

Take the example of 3 distributions $\{p(\vec{x}|y=i)\}_{i=1}^3$ with respective different expectations $\{\mu_i\}_{i=1}^3$. We want to separate them using the MMD OvA GEMINI with linear kernel. The mixture of the 3 distributions creates a data distribution with expectation $\mu=\sum_{i=1}^3 p(y=i)\mu_i$. However if the distributions satisfy that this data expectation $\mu$ is equal to one of the sub-expectations $\mu_i$, then the associated distribution $i$ will be non-evaluated since its MMD to the data distribution is equal to 0. We illustrate this example in figure \ref{fig:example_ova_ovo}. We tackled the problem by introducing the OvO setup where each pair of different cluster distribution is compared.


\section{Another approach to Wasserstein maximisation}
\label{app:wasserstein_maximisation}
The Wasserstein-1 metric can be considered as an \gls*{ipm} defined over a set of 1-Lipschitz functions. Indeed, such writing is the dual representation of the Wasserstein-1 metric:

\begin{equation*}
    W_c(p\|q) = \sup_{f, \|f\|_L\leq 1} \E_{x\sim p}[f(x)] - \E_{z\sim q}[f(z)].
\end{equation*}

Yet, evaluating a supremum as an objective to maximise is hardly compatible with the usual backpropagration in neural networks. This definition was used in attempts to stabilise GAN training \cite{arjovsky_wasserstein_2017} by using 1-Lipschitz neural networks \cite{gouk_regularisation_2021}. However, the Lipschitz continuity was achieved at the time by weight clipping, whereas other methods such as spectral normalisation \cite{miyato_spectral_2018} now allow arbitrarily large weights. The restriction of 1-Lipscthiz functions to 1-Lipschitz neural networks does not equal the true Wasserstein distance, and the term "neural net distance" is sometimes preferred \cite{arora_generalization_2017}. Still, estimating the Wasserstein distance using a set of Lipschitz functions derived from neural networks architectures brings more difficulties to actually leverage the true distance according to the energy cost $c$ of Eq.~\ref{eq:wasserstein_definition}.

Globally, we hardly experimented the generic \gls*{ipm} for \glspl*{gemini}. Our efforts for defining a set of 1-Lipschitz critics, one per cluster for \gls*{ova} or one per pair of clusters for \gls*{ovo}, to perform the neural net distance \cite{arora_generalization_2017} were not fruitful. This is mainly because such objective requires the definition of one neural network for the posterior distribution $\pyx$ and $K$ (resp. $K(K-1)/2$) other 1-Lipschitz neural networks for the \gls*{ova} (resp. \gls*{ovo}) critics, i.e. a large amount of parameters. Moreover, this brings the problem of designing not only one, but many neural networks while the design of one accurate architecture is already a sufficient problem.

\section{Deriving GEMINIs}
\label{app:deriving_geminis}
We show in this appendix how to derive all estimable forms of the \gls*{gemini}.

\subsection{\texorpdfstring{$f$}{f}-divergence GEMINI}

We detail here the derivation for 3 $f$-divergences that we previously chose: the \gls*{kl} divergence, the \gls*{tv} distance and the squared Hellinger distance, as well as the generic scenario for any function $f$.

\subsubsection{Generic scenario}

First, we recall that the definition of an $f$-divergence involves a convex function:

\begin{align*}
    f:\mathbb{R}^+ &\rightarrow \mathbb{R}\\
    x&\rightarrow f(x)\\
    \text{s.t.}\quad f(1)&=0,
\end{align*}

between two distributions $p$ and $q$ as described:

\begin{equation*}
    D_\text{f-div}(p,q) = \E_{\vec{x} \sim q} \left[ f\left(\frac{p(\vec{x})}{q(\vec{x})}\right)\right].
\end{equation*}

We simply inject this definition in the \gls*{ova}-\gls*{gemini} and directly obtain both an expectation on the cluster assignment $y$ and on the data variable $\vec{x}$. We then merge the writing of the two expectations for the sake of clarity.

\begin{align*}
    \mathcal{I}^\text{ova}_\text{f-div} &= \E_{\py} \left[ D_\text{f-div}(\pxy || \px)\right],\\
    &= \E_{\py} \left[ \E_{\px}\left[ f\left(\frac{\pxy}{\px}\right)\right]\right],\\
    &= \E_{\py,\px} \left[ f\left(\frac{\pyx}{\py}\right)\right].\\
\end{align*}

Injecting the $f$-divergence in the \gls*{ovo}-\gls*{gemini} first yields:

\begin{align*}
    \mathcal{I}^\text{ovo}_\text{f-div} &= \E_{\pya,\pyb} \left[ D_\text{f-div}(\pxya || \pxyb)\right],\\
    &= \E_{\pya,\pyb} \left[ \E_{\pxyb}\left[ f\left(\frac{\pxya}{\pxyb}\right)\right]\right].\\
\end{align*}

Now, by using Bayes theorem, we can perform the inner expectation over the data distribution. We then merge the expectations for the sake of clarity.

\begin{align*}
    \mathcal{I}^\text{ovo}_\text{f-div} &= \E_{\pya,\pyb} \left[ \E_{\px} \left[ \frac{\pyxb}{\pyb}f\left(\frac{\pxya}{\pxyb}\right)\right]\right],\\
    &= \E_{\pya,\pyb,\px} \left[ \frac{\pyxb}{\pyb}f\left(\frac{\pyxa\pyb}{\pyxb\pya}\right)\right].\\
\end{align*}

Notice that we also changed the ratio of conditional distributions inside the function by a ratio of posteriors through Bayes' theorem, weighted by the relative cluster proportions.

Now, we can derive into details these equations for the 3 $f$-divergences we focused on: the \gls*{kl} divergence, the \gls*{tv} distance and the squared Hellinger distance.

\subsubsection{Kullback-Leibler divergence}

The function for Kullback-Leibler is $f(t) = t\log t$. We do not need to write the \gls*{ova} equation: it is straightforwardly the usual \gls*{mi}. For the \gls*{ovo}, we inject the function definition by replacing:

\begin{equation*}
    t=\frac{\pyxa\pyb}{\pyxb\pya},
\end{equation*}

in order to get:

\begin{align*}
    \mathcal{I}^\text{ovo}_\text{KL} &=  \E_{\pya,\pyb,\px} \left[ \frac{\pyxb}{\pyb}\times\frac{\pyxa\pyb}{\pyxb\pya}\log{\frac{\pyxa\pyb}{\pyxb\pya}}\right].
\end{align*}

We can first simplify the factors outside of the logs:

\begin{align*}
    \mathcal{I}^\text{ovo}_\text{KL}&=\E_{\pya,\pyb,\px} \left[ \frac{\pyxa}{\pya}\log{\frac{\pyxa\pyb}{\pyxb\pya}}\right].
\end{align*}

If we use the properties of the log, we can separate the inner term in two sub-expressions:

\begin{align*}
    \mathcal{I}^\text{ovo}_\text{KL} = \E_{\pya,\pyb,\px} \left[ \frac{\pyxa}{\pya}\log{\frac{\pyxa}{\pya}} + \frac{\pyxa}{\pya}\log{\frac{\pyb}{\pyxb}}\right].
\end{align*}

Hence, we can use the linearity of the expectation to separate the two terms above. The first term is constant w.r.t. $\yb$, so we can remove this variable from the expectation among the subscripts:

\begin{multline*}
    \mathcal{I}^\text{ovo}_\text{KL} = \E_{\pya,\px} \left[ \frac{\pyxa}{\pya}\log{\frac{\pyxa}{\pya}} \right] + \E_{\pya,\pyb,\px} \left[ \frac{\pyxa}{\pya}\log{\frac{\pyb}{\pyxb}} \right].
\end{multline*}

Since the variables $\ya$ and $\yb$ are independent, we can use the fact that:

\begin{equation*}
    \E_{\pya}\left[\frac{\pyxa}{\pya}\right] = \int \pya \frac{\pyxa}{\pya} d\ya = 1,
\end{equation*}

inside the second term to reveal the final form of the equation:

\begin{align*}
    \mathcal{I}^\text{ovo}_\text{KL} = \E_{\px,\py} \left[ \frac{\pyx}{\py}\log{\frac{\pyx}{\py}} \right] + \E_{\px,\py} \left[ \log{\frac{\py}{\pyx}}\right].
\end{align*}

Notice that since both terms did not compare one cluster assignment $\ya$ against another $\yb$, we can switch to the same common variable $y$. Both terms are in fact \gls*{kl} divergences depending on the cluster assignment $y$. The first is the reverse of the second. This sum of \gls*{kl} divergences is sometimes called the \emph{symmetric} \gls*{kl}, and so can we write in two ways the \gls*{kl}-\gls*{ovo} \gls*{gemini}:

\begin{align*}
    \mathcal{I}^\text{ovo}_\text{KL} &= \E_{\px} \left[ D_\text{KL} (\pyx || \py)\right] + \E_{\px} \left[ D_\text{KL} (\py || \pyx)\right],\\
    &= \E_{\px} \left[ D_\text{KL-sym}(\pyx || \py)\right].\\
\end{align*}

We can also think of this equation as the usual \gls*{mi} with an additional term based on the reversed \gls*{kl} divergence.

\subsubsection{Total Variation distance}

For the total variation, the function is $f(t)=\frac{1}{2} |t-1|$. Thus, the \gls*{ova} \gls*{gemini} is:

\begin{equation*}
    \mathcal{I}^\text{ova}_\text{TV} = \frac{1}{2}\E_{\py,\px} \left[ |\frac{\pyx}{\py}-1|\right].
\end{equation*}

And the \gls*{ovo} is:

\begin{align*}
    \mathcal{I}^\text{ovo}_\text{TV} &=\frac{1}{2}\E_{\pya,\pyb,\px} \left[ \frac{\pyxb}{\pyb}|\frac{\pyxa\pyb}{\pyxb\pya}-1|\right],\\
    &=\frac{1}{2}\E_{\pya,\pyb,\px} \left[ |\frac{\pyxa}{\pya} - \frac{\pyxb}{\pyb} | \right].
\end{align*}

We did not find any further simplification of these equations.

\subsubsection{Squared Hellinger distance}

Finally, the squared Hellinger distance is based on $f(t)=2(1-\sqrt{t})$. Hence the \gls*{ova} unfolds as:

\begin{align*}
    \mathcal{I}^\text{ova}_{\text{H}^2}&= \E_{\py,\px} \left[ 2\left(1-\sqrt{\frac{\pyx}{\py}}\right)\right],\\
    &= 2-2\E_{\px,\py} \left[\sqrt{\frac{\pyx}{\py}}\right].
\end{align*}

The idea of the squared Hellinger-\gls*{ova} \gls*{gemini} is therefore to minimise the expected square root of the relative certainty between the posterior and cluster proportion.

For the \gls*{ovo} setting, the definition yields:

\begin{equation*}
    \mathcal{I}^\text{ovo}_{\text{H}^2}=\E_{\pya,\pyb,\px} \left[ \frac{\pyxb}{\pyb}\times 2 \times\left(1-\sqrt{\frac{\pyxa\pyb}{\pyxb\pya}}\right)\right],
\end{equation*}

which we can already simplify by putting the constant 2 outside of the expectation, and by inserting all factors inside the square root before simplifying and separating the expectation:

\begin{align*}
    \mathcal{I}^\text{ovo}_{\text{H}^2} &= 2\E_{\pya,\pyb,\px} \left[ \frac{\pyxb}{\pyb} - \frac{\pyxb}{\pyb}\sqrt{\frac{\pyxa\pyb}{\pya\pyxb}}\right],\\
    &= 2\E_{\pya,\pyb,\px} \left[ \frac{\pyxb}{\pyb} \right] - 2\E_{\pya,\pyb,\px} \left[ \sqrt{\frac{\pyxa \pyxb}{\pya \pyb}}\right].
\end{align*}

We can replace the first term by the constant 1, as shown for the \gls*{kl}-\gls*{ovo} derivation. Since we can split the square root into the product of two square roots, we can apply twice the expectation over $\ya$ and $\yb$ because these variables are independent:

\begin{equation*}
    \mathcal{I}^\text{ovo}_{\text{H}^2} = 2-2\E_{\px} \left[ \E_{\py} \left[\sqrt{\frac{\pyx}{\py}}\right]^2\right].
\end{equation*}

To avoid computing this squared expectation, we use the equation of the variance $\mathbb{V}$ to replace it. Thus:

\begin{align*}
    \mathcal{I}^\text{ovo}_{\text{H}^2} &= 2-2\E_{\px} \left[ \E_{\py}\left[\frac{\pyx}{\py}\right] - \mathbb{V}_{\py}\left[\sqrt{\frac{\pyx}{\py}}\right]\right],\\
    &= 2 - 2\E_{\px}\left[ \E_{\py} \left[ \frac{\pyx}{\py}\right]\right] + 2 \E_{\px} \left[ \mathbb{V}_{\py} \left[ \sqrt{\frac{\pyx}{\py}}\right]\right].
\end{align*}

Then, for the same reason as before, the second term is worth 1, which cancels the first constant. We therefore end up with:

\begin{equation*}
    \mathcal{I}^\text{ovo}_{\text{H}^2}= 2\E_{\px} \left[ \mathbb{V}_{\py}\left[\sqrt{\frac{\pyx}{\py}}\right]\right].
\end{equation*}

Similar to the \gls*{kl}-\gls*{ovo} case, the squared Hellinger \gls*{ovo} converges to an \gls*{ova} setting, i.e. we only need information about the cluster distribution itself without comparing it to another. Furthermore, the idea of maximising the variance of the cluster assignments is straightforward for clustering.

\subsection{Maximum Mean Discrepancy}

When using an IPM with a family of functions that project an input of $\mathcal{X}$ to the unit ball of an RKHS $\mathcal{H}$, the \gls*{ipm} becomes the \gls*{mmd} distance.

\begin{align*}
\text{MMD}(p,q) &= \sup_{f: ||f||_\mathcal{H}\leq 1} \E_{\xa\sim p}[f(\xa)] - \E_{\xb \sim q} [f(\xb)],\\
&= \| \E_{\xa\sim p} [\varphi(\xa)] - \E_{\xb \sim q}[\varphi(\xb)]\|_{\mathcal{H}},\\
\end{align*}

where $\varphi$ is a embedding function of the RKHS.

By using a kernel function $\kappa(\xa,\xb) = <\varphi(\xa), \varphi(\xb)>$, we can express the square of this distance thanks to inner product space properties \cite{gretton_kernel_2012}:

\begin{align*}
\text{MMD}^2 (p,q) &= \E_{\xa, \xa^\prime \sim p}[\kappa(\xa, \xa^\prime)] + \E_{\xb,\xb^\prime \sim q}[\kappa(\xb, \xb^\prime)] - 2 \E_{\xa\sim p, \xb\sim q}[\kappa(\xa, \xb)].
\end{align*}


Now, we can derive each term of this equation using our distributions $p\equiv\pxy$ and $q\equiv \px$ for the \gls*{ova} case, and $p\equiv\pxya, q\equiv\pxyb$ for the \gls*{ovo} case. In both scenarios, we aim at finding an expectation over the data variable $x$ using only the respectively known and estimable terms $\pyx$ and $\py$.

\paragraph{OvA scenario}

For the first term, we use Bayes' theorem twice to get an expectation over two variables $\xa$ and $\xb$ drawn from the data distribution.

\begin{equation*}
    \begin{split}
        \E_{\xa, \xa^\prime \sim p}&= \E_{\xa, \xa^\prime \sim \pxy} \left[ \kappa(\xa,\xa^\prime)\right],\\
        &= \E_{\xa,\xa^\prime \sim \px} \left[ \frac{\p(y|\xa)\p(y|\xa^\prime)}{\py^2} \kappa(\xa,\xa^\prime)\right].
    \end{split}
\end{equation*}

For the second term, we do not need to perform anything particular as we directy get an expectation over the data variabes $\xa$ and $\xb$.

\begin{equation*}
    \E_{\xb, \xb^\prime \sim q}= \E_{\xb, \xb^\prime \sim \px} \left[ \kappa(\xb,\xb^\prime)\right].
\end{equation*}

The last term only needs Bayes theorem once, for the distribution $q$ is directly replaced by the data distribution $\px$:

\begin{align*}
    \E_{\xa\sim p, \xb \sim q}&=\E_{\xa \sim \pxy, \xb \sim \px}\left[ \kappa(\xa,\xb)\right],\\
    &= \E_{\xa, \xb \sim px} \left[ \frac{\p(y|\xa)}{\py} \kappa(\xa,\xb)\right].
\end{align*}

Note that for the last term, we could replace $\p(y|\xa)$ by $\p(y|\xb)$; that would not affect the result since $\xa$ and $\xb$ are independently drawn from $\px$. We thus replace all terms, and do not forget to put a square root on the entire sum since we have computed so far the squared \gls*{mmd}:

\begin{align*}
        \mathcal{I}_\text{MMD}^\text{ova} &= \E_{\py} \left[ \text{MMD}(\pxy,\px)\right],\\
        &=\E_{\py} \left[ \left(\E_{\xa,\xa^\prime \sim \px} \left[ \frac{\p(y|\xa)\p(y|\xa^\prime)}{\py^2} \kappa(\xa,\xa^\prime)\right] \right.\right.\\
        &\left.\left.\qquad+ \E_{\xb, \xb^\prime \sim \px} \left[ \kappa(\xb,\xb^\prime)\right] -2 \E_{\xa, \xb \sim \px} \left[ \frac{\p(y|\xa)}{\py} \kappa(\xa,\xb)\right] \right)^{\frac{1}{2}}  \right].\\
\end{align*}

Since all variables $\xa$, $\xa^\prime$, $\xb$ and $\xb^\prime$ are independently drawn from the same distribution $\px$, we can replace all of them by the variables $\vec{x}$ and $\vec{x}^\prime$. We then use the linearity of the expectation and factorise by the kernel $\kappa(\vec{x},\vec{x}^\prime)$:

\begin{equation*}
        \mathcal{I}_\text{MMD}^\text{ova} = \E_{\py} \left[ \E_{\vec{x},\vec{x}^\prime \sim \px}\left[ \kappa(\vec{x},\vec{x}^\prime) \left( \frac{\p(y|\vec{x})\p(y|\vec{x}^\prime)}{\py^2} + 1 - 2\frac{\p(y|\vec{x})}{\py}\right) \right]^{\frac{1}{2}}\right].
\end{equation*}

\paragraph{OvO scenario}

The two first terms of the \gls*{ovo} \gls*{mmd} are the same as the first term of the \gls*{ova} setting, with a simple subscript $a$ or $b$ at the appropriate place. Only the negative term changes. We once again use Bayes' theorem twice:

\begin{align*}
        \E_{\xa \sim p, \xb \sim q} [\kappa(\xa, \xb)]&= \E_{\xa \sim \pxya, \xb \pxyb} \left[ \kappa(\xa,\xb) \right],\\
        &= \E_{\xa, \xb \sim \px} \left[\frac{\p(\ya|\xa)}{\pya}\frac{\p(\yb|\xb)}{\pyb} \kappa(\xa,\xb) \right].
\end{align*}

The final sum is hence similar to the \gls*{ova}:

\begin{align*}
        \mathcal{I}_\text{MMD}^\text{ovo} &= \E_{\pya, \pyb} \left[ \text{MMD} (\pxya, \pxyb)\right],\\
        &= \E_{\pya, \pyb} \left[ \E_{\vec{x},\vec{x}^\prime \sim \px} \left[ \kappa(\vec{x},\vec{x}^\prime) \left( \frac{\pyxa \p(\ya|\vec{x}^\prime)}{\pya^2} + \frac{\pyxb \p(\yb|\vec{x}^\prime)}{\pyb^2}\right.\right.\right.\\&\quad\left.\left.\left.-2 \frac{\pyxa \p(\yb|\vec{x}^\prime) }{\pya \pyb} \right) \right]^{\frac{1}{2}}\right].
\end{align*}

\subsection{Wasserstein distance}
\label{app:wasserstein_convergence}

To compute the Wasserstein distance between the distributions $\p(\vec{x}|y=k)$, we estimate it using approximate distributions. We replace $\p(\vec{x}|y=k)$ by a weighted sum of Dirac measures on specific samples $\vec{x}_i$: $p_N^k$:

\begin{equation*}
    \p(\vec{x}|y=k) \approx \sum_{i=1}^N m_i^k\delta_{\vec{x}_i} = p_N^k,
\end{equation*}

where $\{m_i^k\}_{i=1}^N$ is the set of weights. We now show that computing the Wasserstein distance between these approximates converges to the correct distance. We first need to show that $p_N^k$ weakly converges to $p$. To that end, let $f$ be any bounded and continuous function. Computing the expectation of such through $\p$ is:

\begin{equation*}
\E_{\vec{x}\sim \p(\vec{x}|y=k)}[f(\vec{x})] = \int_{\mathcal{X}}f(\vec{x})\p(\vec{x}|y=k)d\vec{x},
\end{equation*}

which can be estimated using self-normalised importance sampling~\cite[Chapter 9]{owen_monte_2009}. The proposal distribution we take for sampling is $\px$. Although we cannot evaluate both $\pxy$ and $\px$ up to a constant, we can evaluate their ratio up to a constant which is sufficient:

\begin{align*}
\E_{\vec{x}\sim \p(\vec{x}|y=k)}[f(\vec{x})]&= \int_{\mathcal{X}}f(\vec{x})\frac{\p(\vec{x}|y=k)}{\px}\px d\vec{x},\\
&= \int_{\mathcal{X}}f(\vec{x})\frac{\p(y=k|\vec{x})}{\p(y=k)}\px d\vec{x},\\
&\approx \sum_{i=1}^N f(\vec{x}_i) \frac{\p(y=k|\vec{x}=\vec{x}_i)}{\sum_{j=1}^N \p(y=k|\vec{x}=\vec{x}_j)}.
\end{align*}

Now, by noticing in the last line that the importance weights are self normalised and add up to 1, we can identify them as the point masses of our previous Dirac approximations:

\begin{equation*}
m_i^k = \frac{\p(y=k|\vec{x}=\vec{x}_i)}{\sum_{j=1}^N \p(y=k|\vec{x}=\vec{x}_j)}.
\end{equation*}

This allows to write that the Monte Carlo estimation through importance sampling of the expectation w.r.t $\p(\vec{x}|y=k)$ is directly the expectation taken on the discrete approximation $p_N^k$. We can conclude that there is a convergence between the two expectations owing to the law of large numbers:

\begin{equation*}
\lim_{N\rightarrow +\infty}\E_{\vec{x}\sim p_N^k}[f(\vec{x})]  = \E_{\vec{x}\sim \p(\vec{x}|y=k)}[f(\vec{x})].
\end{equation*}

Since $f$ is bounded and continuous, the portmanteau theorem~\cite{billingsley_convergence_2013} states that $p_N^k$ weakly converges to $\p(\vec{x}|y=k)$ when defining the importance weights as the normalised predictions cluster-wise.

To conclude, when two series of measures $p_N$ and $q_N$ weakly converge respectively to $p$ and $q$, so does their Wasserstein distance ~\cite[Corollary 6.9]{villani_optimal_2009}, hence:

\begin{equation}
    \lim_{N\rightarrow+\infty}\mathcal{W}_c(p_N^{k_1},p_N^{k_2})= \mathcal{W}_c\left(\p(\vec{x}|y=k_1)\|\p(\vec{x}|y=k_2)\right).
\end{equation}

\section{More information and experiment on the Gaussian and Student-t distributions mixture}
\label{app:gstm}
\subsection{Generative process of Gaussian and Student Mixture}

We describe here the generative protocol for the Gaussian and Student mixture dataset. Each cluster distribution is centered around a mean $\mu_i$ which proximity is controlled by a scalar $\alpha$. For simplicity, all covariance matrices are the identity scaled by a scalar $\sigma$. We define:

\begin{equation*}
    \mu_1 = [\alpha, \alpha],\quad\mu_2 = [\alpha, -\alpha],\quad\mu_3 = [-\alpha, \alpha],\quad\mu_4 = [-\alpha, -\alpha].
\end{equation*}

To sample from a multivariate Student-t distribution, we first draw samples $x$ from a centered multivariate Gaussian distribution. We then sample another variable $u$ from a $\chi^2$-distribution using the degrees of freedom $\rho$ as parameter. Finally, $x$ is multiplied by $\sqrt{\frac{\rho}{u}}$, yielding samples from the Student-t distribution.

\subsection{Extended experiment}

\begin{table}
    \caption{Mean ARI (std) of a MLP fitting a mixture of 3 Gaussian and 1 Student-t multivariate distributions compared with Gaussian Mixture Models and K-Means. The model can be tasked to find either 4 or 8 clusters at best and the Student-t distribution has $\rho$=1 or 2 degrees of freedom. Bottom line presents the ARI for the maximum a posteriori of an oracle aware of all parameters of the data.}
    \label{tab:complete_gstm_exp}
    \centering
    \begin{tabular}{c c c c c}
        \toprule
        \multirow{2}[3]{*}{Model}& \multicolumn{2}{c}{$\rho=2$}&\multicolumn{2}{c}{$\rho=1$}\\
        \cmidrule(lr){2-3}\cmidrule(lr){4-5}
        & 4 clusters&8 clusters&4 clusters&8 clusters\\
        \midrule
        K-Means&0.965 (0)&0.897 (0.040)&0 (0)&0.657 (0.008)\\
        GMM (full covariance)&0.972 (0)&0.868 (0.042)&0 (0)&0.610 (0.117)\\
        GMM (diagonal covariance)&{\bf 0.973 (0)}& 0.862 (0.048)&0.024 (0.107)& 0.660 (0.097)\\
        \midrule
        $\I_\text{KL}^\text{ova}$&0.883 (0.182)&0.761 (0.101)&{\bf 0.939 (0.006)}&0.742 (0.092)\\
        $\I_\text{KL}^\text{ovo}$&0.731 (0.140)&0.891 (0.129)&0.723 (0.114)&0.755 (0.163)\\
        $\I_{\text{H}^2}^\text{ova}$&0.923 (0.125)&{\bf 0.959 (0.043)}&0.906 (0.103)&0.86 (0.087)\\
        $\I_{\text{H}^2}^\text{ovo}$&0.926 (0.112)&0.951 (0.059)&0.858 (0.143)&0.887 (0.074)\\
        $\I_\text{TV}^\text{ova}$&0.940 (0.097)&0.973 (0.004)&0.904 (0.104)&{\bf 0.925 (0.103)}\\
        $\I_\text{TV}^\text{ovo}$&0.971 (0.005)&0.620 (0.053)&{\bf 0.938 (0.005)}&0.595 (0.055)\\
        \midrule
        $\I_\text{MMD}^\text{ova}$&0.953 (0.060)&0.940 (0.033)&0.922 (0.004)&{\bf 0.908 (0.016)}\\
        $\I_\text{MMD}^\text{ovo}$&0.968 (0.001)&0.771 (0.071)&0.921 (0.007)&0.849 (0.048)\\
        $\I_\mathcal{W}^\text{ova}$&0.897 (0.096)&0.896 (0.021)&0.915 (0.131)&0.889 (0.051)\\
        $\I_\mathcal{W}^\text{ovo}$&0.970 (0.002)&0.803 (0.067)&0.922 (0.006)&0.817 (0.042)\\
        \midrule
        Oracle&\multicolumn{2}{c}{0.991}&\multicolumn{2}{c}{0.989}\\
        \bottomrule
    \end{tabular}
\end{table}

For the main experiment, we fixed $\sigma=1$, $\alpha=5$ and the degree of freedom $\rho=1$. We further tested our method when training a \gls*{mlp} for 4 or 8 clusters. For both cases, we also considered $\rho=1$ and $\rho=2$. We include the complete results in Table~\ref{tab:complete_gstm_exp}.

\section{Model selection on MNIST}
\label{app:cluster_selection}
\begin{figure}
\centering
    \subfloat[MLP]{
        \includegraphics[width=0.45\linewidth]{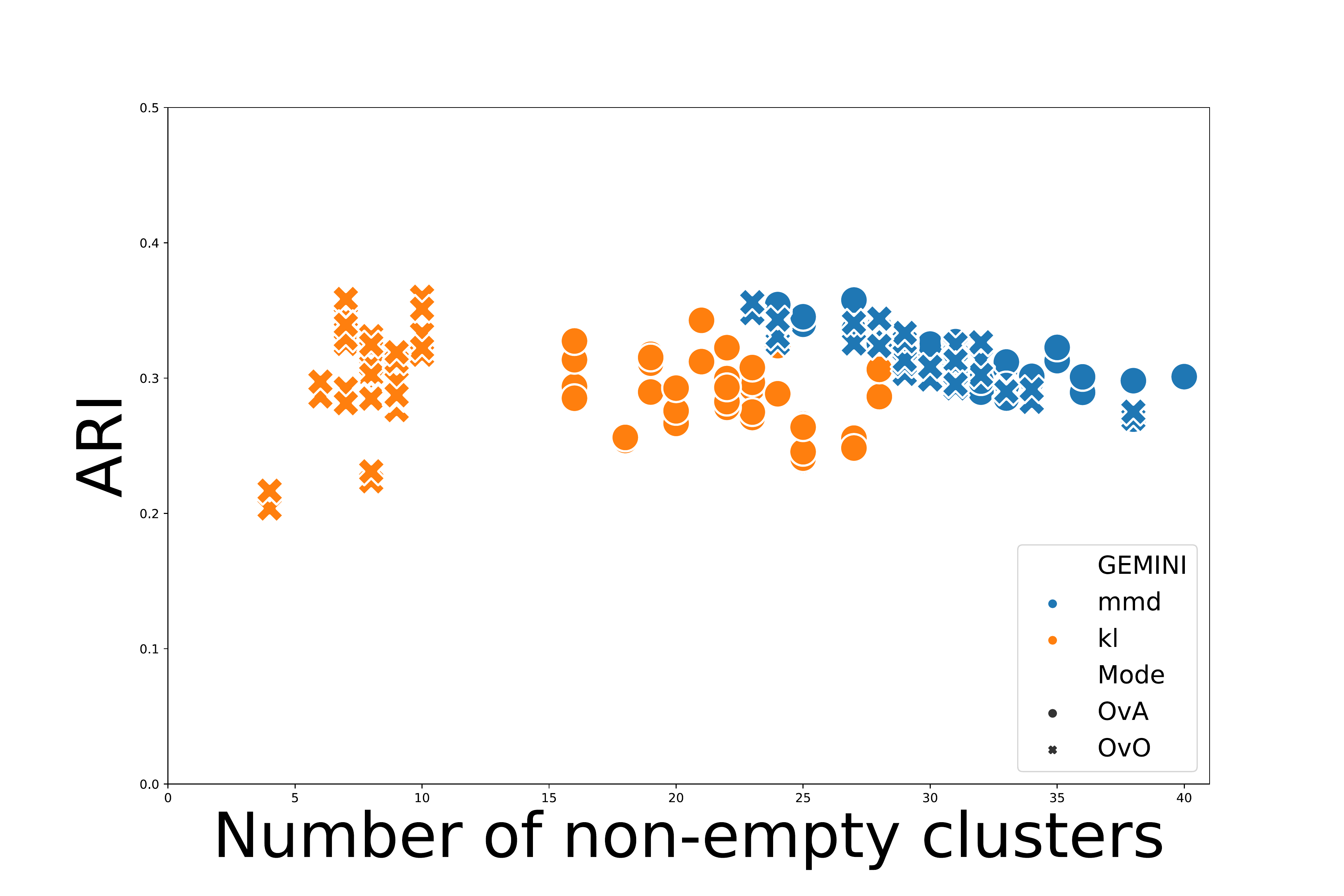}
        \label{sfig:mlp_clusters}
    }\hfill
    \subfloat[LeNet5]{
        \includegraphics[width=0.45\linewidth]{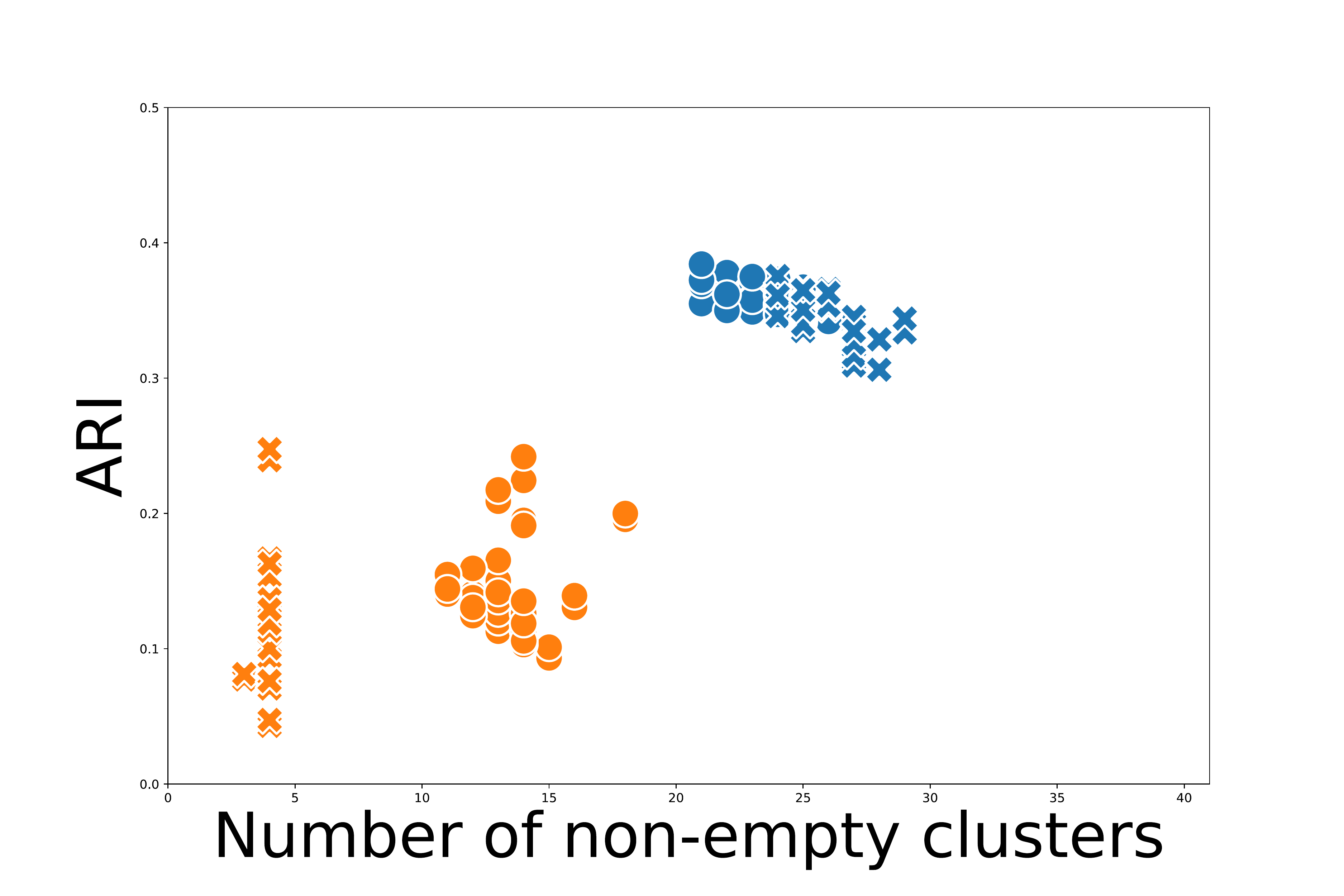}
        \label{sfig:lenet_clusters}
    }
    \caption{Distributions of the ARI scores given a number of non-empty clusters after 100 epochs of training on MNIST on two different architectures.}
    \label{fig:mnist_clusters}
\end{figure}

We repeat our previous experiment on the MNIST dataset from Sec.~\ref{ssec:exp_mnist}. We choose this time to get 50 clusters at best for both the MI and the MMD GEMINI and train the models for 100 epochs. We repeat the experiment 20 times per model and plot the resulting scores in figures~\ref{sfig:mlp_clusters} and~\ref{sfig:lenet_clusters}. We did not choose to test with the Wasserstein GEMINI because its complexity implies a long training time for 50 clusters, as explained in App.~\ref{app:exp_complexity}. We first observe in Fig.~\ref{fig:mnist_clusters} that the MMD-GEMINI with linear kernel has a tendency to exploit more clusters than the MI. The model converges to approximately 30 clusters in the case of the MLP and 25 for the LeNet-5 model with less variance. We can further observe that for all metrics the choice of architecture impacted the number of non-empty clusters after training. Indeed, by playing a key role in the decision boundary shape, the architecture may limit the number of clusters to be found: the MLP can draw more complex boundaries compared to the LeNet5 model. Moreover, we suppose that the cluster selection behaviour of GEMINI may be due to optimisation processes. Indeed, we optimise estimators of the GEMINI rather than the exact GEMINI.
Finally, Fig.~\ref{fig:mnist_clusters} also confirms from Table.~\ref{tab:mnist_experiment} the stability of the MMD-GEMINI regarding the ARI despite the change of architecture whereas the MI is affected and shows poor performance with the LeNet-5 architecture.

\section{Choosing a GEMINI}
\label{app:exp_complexity}
\begin{figure}[hbt]
    \centering
    \subfloat[10 samples per batch]{
        \includegraphics[width=0.45\linewidth]{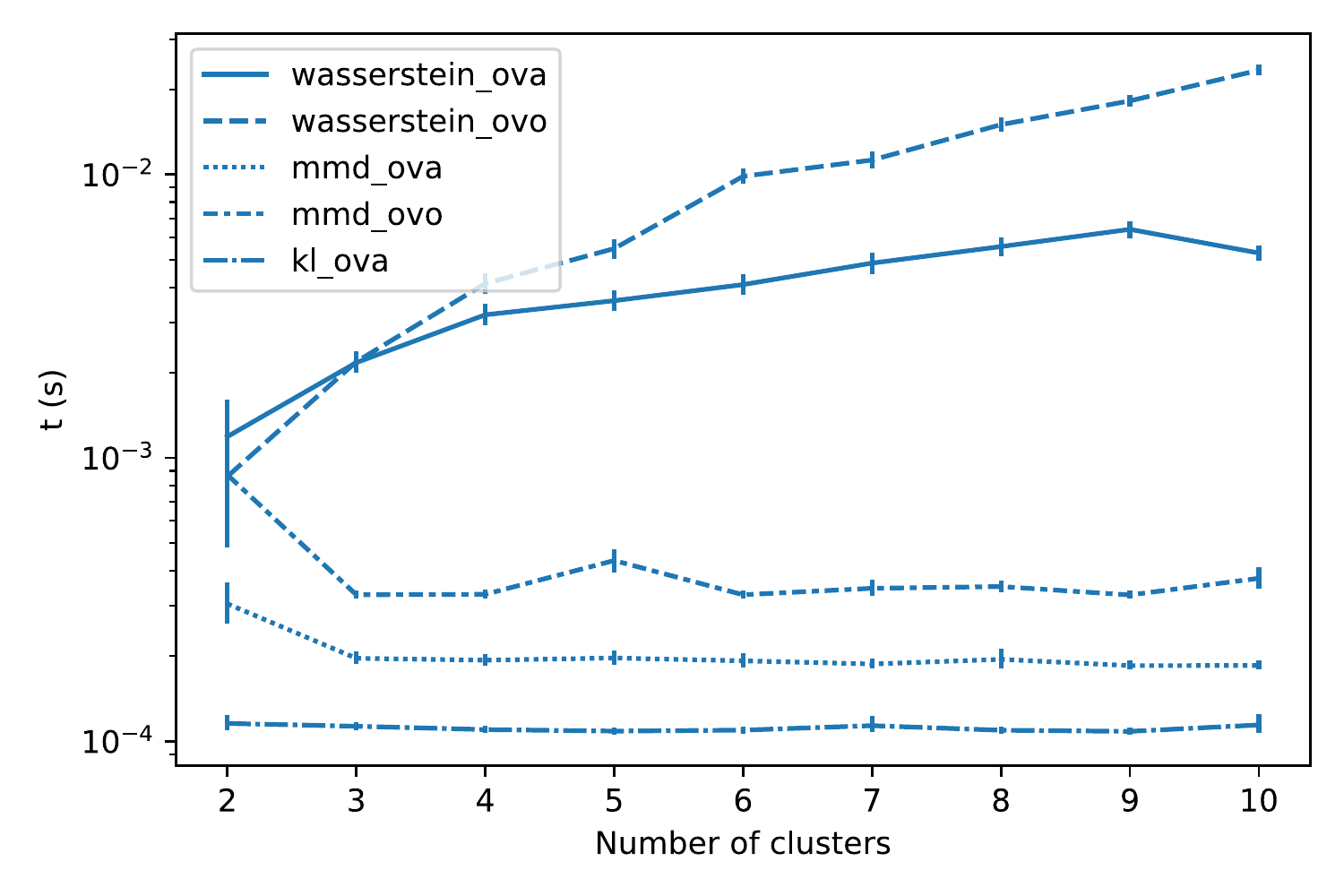}
        \label{sfig:time_preformances_batch10}
    }
    \subfloat[500 samples per batch]{
        \includegraphics[width=0.45\linewidth]{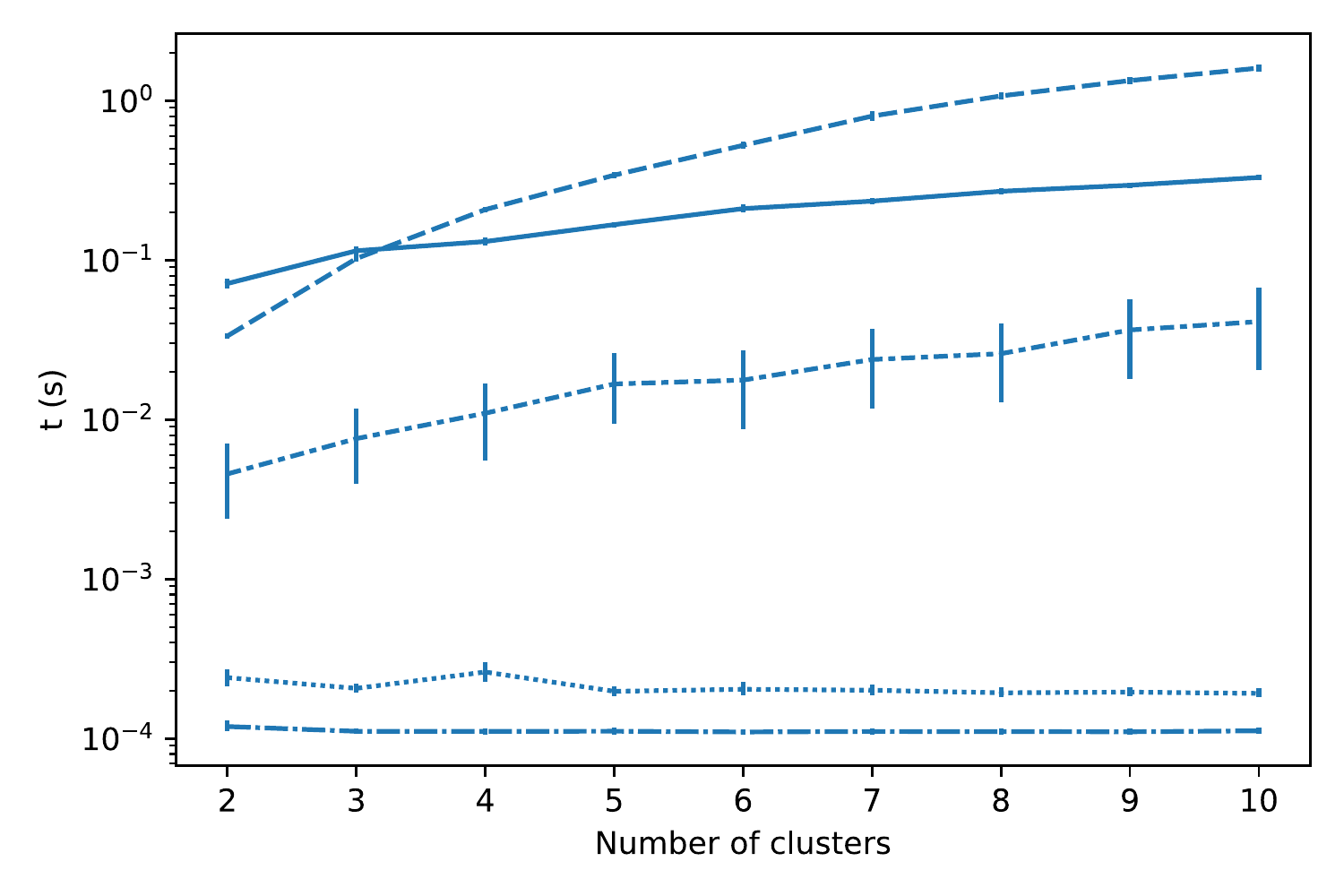}
        \label{sfig:time_performances_batch500}
    }\hfill
    \caption{Average performance time (in seconds) of GEMINIs as the number of tasked clusters grows for batches of size 10 and 500 samples.}
    \label{fig:time_performances}
\end{figure}

The complexity of \gls*{gemini} increases with the distances previously mentionned depending on the number of clusters $K$ and the number of samples per batch $N$. It ranges from $\mathcal{O}(NK)$ for the usual \gls*{mi} to $\mathcal{O}(K^2N^3\log{N})$ for the Wasserstein-\gls*{gemini}-\gls*{ovo}. As an example, we show in Figure~\ref{fig:time_performances} the average time of \gls*{gemini} as the number of tasked clusters increases for both 10 samples per batch (Figure~\ref{sfig:time_preformances_batch10}) and 500 samples (Figure~\ref{sfig:time_performances_batch500}). The batches consists in randomly generated prediction and distances or kernel between randomly generated data.

The Wasserstein-\gls*{ovo} is the most complex, and so its usage should remain for 10 clusters or less overall. The second most time-consuming loss is the Wasserstein-\gls*{ova}, however its tendency in optimisation to only find 2 clusters makes it a inappropriate. The main difference also to notice between the following \gls*{mmd} is regarding their memory complexity. The \gls*{mmd}-\gls*{ova} requires only $\mathcal{O}(KN^2)$ while the \gls*{mmd}-\gls*{ovo} requires $\mathcal{O}(K^2N^2)$. This memory complexity should be the major guide to choosing one \gls*{mmd}-\gls*{gemini} or the other. Thus, the minimal time-consuming and resource-demanding \gls*{gemini} is the \gls*{mmd}-\gls*{ova} if we consider \glspl*{gemini} that incorporates knowledge of data through kernels and distances. Other versions involving $f$-divergences have in fact the same complexity as \gls*{mi} in our implementations, apart from the \gls*{tv}-\gls*{ovo} which reaches $\mathcal{O}(K^2N)$ in our implementation.

\section{All pair shortest paths distance}
\label{app:fw_distance}
Sometimes, using distances such as the $\ell_2$ may not capture well the shape of manifolds. To do so, we derive a metric using the all pair shortest paths. Simply put, this metric consists in considering the number of closest neighbors that separates two data samples. To compute it, we first use a sub-metric that we note $d$, say the $\ell_2$ norm. This allows us to compute all distances $d_{ij}$ between every sample $i$ and $j$. From this matrix of sub-distances, we can build a graph adjacency matrix $W$ following the rules:

\begin{equation}
    W_{ij} = \left\{\begin{array}{cr}
        1 & d_{ij}\leq \epsilon \\
        0 & d_{ij}> \epsilon
    \end{array}\right.,
\end{equation}

where $\epsilon$ is a chosen threshold such that the graph has sparse edges. Our typical choice for $\epsilon$ is the 5\% quantile of all $d_{ij}$.

We chose the graph adjacency matrix to be undirected, owing to the symmetry of $d_{ij}$ and unweighted. Indeed, solving the all-pairs shortest paths involves the Floyd-Warshall algorithm \cite{warshall_theorem_1962,roy_transitivite_1959} which complexity $\mathcal{O}(n^3)$ is not affordable when the number of samples $n$ becomes large. An undirected and unweighted graph leverages performing $n$ times the breadth-first-search algorithm, yielding a total complexity of $\mathcal{O}(n^2+ne)$ where $e$ is the number of edges. Consequently, setting a good threshold $\epsilon$ controls the complexity of the shortest paths to finds. Our final distance between two nodes $i$ and $j$ is eventually:

\begin{equation}
    c_{ij} = \left\{\begin{array}{c r}\text{Shortest-path}^W(i,j)&\text{if it exists.}\\
    n&\text{otherwise}
    \end{array}\right..
\end{equation}

This metric $c$ can then be incorporated inside the Wasserstein-\gls*{gemini}.

\section{Packages for experiments}
\label{app:requirements}
For the implementation details, we use several packages with a python 3.8 version.
\begin{itemize}
    \item We use PyTorch \cite{pytorch} for all deep learning models and automatic differentiation, as well as NumPy \cite{numpy} for arrays handling.
    \item We use Python Optimal Transport's function \verb+emd2+\cite{flamary_pot_2021} to compute the Wasserstein distances between weighted sums of Diracs.
    \item We used the implementation of SIMCLR from PyTorch Lightning \cite{pytorch_lightning}.
    \item Small datasets such as isotropic Gaussian Mixture of score computations are performed using scikit-learn\cite{scikit-learn}.
    \item All figures were generated using Pyplot from matplotlib \cite{matplotlib}.
\end{itemize}

\end{document}